\documentclass{article}

\usepackage{subcaption}
\usepackage[table]{xcolor}
\usepackage{hyperref}
\usepackage{enumitem}
\usepackage{arxiv}

\usepackage[utf8]{inputenc} % allow utf-8 input
\usepackage[T1]{fontenc}    % use 8-bit T1 fonts
\usepackage{hyperref}       % hyperlinks
\usepackage{url}            % simple URL typesetting
\usepackage{booktabs}       % professional-quality tables
\usepackage{amsfonts}       % blackboard math symbols
\usepackage{nicefrac}       % compact symbols for 1/2, etc.
\usepackage{microtype}      % microtypography
\usepackage{lipsum}
\usepackage{graphicx}
\usepackage{amssymb}
\title{Filter design for small target detection on infrared imagery using normalized-cross-correlation layer}
\author{
\textbf{H. Se\c{c}kin DEM\.IR$^{1}$, Erdem AKAG\"UND\"UZ$^{2}$}\\
$^{1}$Department of Electro-Optics Systems Design, MGEO, ASELSAN Inc., Turkey\\
$^{2}$Department of Electrical and Electronics Engineering, \c{C}ankaya University, Turkey}

\begin{document}
\maketitle

\begin{abstract}In this paper, we introduce a machine learning approach to the problem of infrared small target detection filter design. For this purpose, similarly to a convolutional layer of a neural network, the normalized-cross-correlational (NCC) layer, which we utilize for designing a target detection/recognition filter bank, is proposed. By employing the NCC layer in a neural network structure, we introduce a framework, in which supervised training is used to calculate the optimal filter shape and the optimum number of filters required for a specific target detection/recognition task on infrared images. We also propose the mean-absolute-deviation NCC (MAD-NCC) layer, an efficient implementation of the proposed NCC layer, designed especially for FPGA systems, in which square root operations are avoided for real-time computation. As a case study we work on dim-target detection on mid-wave infrared imagery and obtain the filters that can discriminate a dim target from various types of background clutter, specific to our operational concept.

\keywords{Small Target Detection, Filter Design, Normalized-Cross Correlation, Convolutional Neural Networks
}
\end{abstract}

\section{Introduction}
\label{Int}
{S}{mall} target detection on infrared (IR) imagery is one of the basic, yet challenging problems of infrared vision. Depending on the detection range, target size, clutter level, operational success requirements, computation power availability and power constraints, several solutions \cite{Zhang2003, BAI2011,Sanna2014} have been proposed for various systems, such as infrared search-and-track (IRST), forward-looking infrared (FLIR), just to name a few.  Although there are countless approaches to the problem, an efficient, widely-accepted and off-the-shelf solution still does not exist for small target detection or recognition problem on infrared imagery. 

Conventional solutions on small target detection on IR imagery \cite{Warren2002,Barnett1989,Tom1993,Nitzberg1979,Schmidt1990,Wen2003,yang2004adaptive,bai2010analysis,bae2012edge,deng2016small,wang2012infrared,kim2012scale,gao2013infrared,chen2014local,han2016infrared,wei2016multiscale} aim at reducing the background clutter by proposing different filter types in a heuristic manner, such as mean, median, top-hat, Gaussian etc. Some of these methods are specifically designed for infrared small target detection \cite{gao2013infrared,chen2014local,han2014robust,bai2010analysis,kim2012scale,deng2016small}. Although these filters show some success to reduce the clutter, they are not inherently designed to detect a specific type of target. Or they do not have the ability to differentiate a target from a false target, which is usually not clutter, but a different object in the scene, like a bird or a bad pixel. Multi-scale approaches \cite{Gregoris1994,Wang1997,Wang2006,Kim2012,Wei2016} to the problem provide scale-invariance; thus they are robust to target size change. However, neither the multi-scale approaches nor some recent entropy \cite{Deng2017} or saliency-based \cite{Qi2016} methods promise sufficient performance against false targets or target-like clutter.  

Using correlation filters to detect small or extended targets in a signal is a well-studied approach \cite{Alam2014}. Especially for different infrared imaging sub-bands such as mid-wave (MW) or long-wave (LW) infrared, normalized-cross-correlation (NCC) is proven to be an effective template matching method \cite{Gundogdu2015}. However, the problem with the NCC-based matching is the ambiguity in filter selection or design. To solve this problem, the idea of supervised filter training is previously introduced, in which the required filter is designed using a dataset \cite{Shao2008}. Especially to solve the tracking problem where the target-clutter relation constantly varies, learning-based approaches are highly effective \cite{Gundogdu2017,Valmadre2017}.

In this paper, we introduce a learning-based approach to small target detection filter design problem on infrared imagery. To this end, we propose the \emph{normalized-cross-correlational} layer, a variation of convolutional layers of a neural network. Utilizing the NCC layer, we introduce a framework, in which supervised training is used to compute a filter bank, i.e. the optimal filters and the optimum number of filters required for a specific detection/recognition task. By feeding the proposed normalized-cross-correlational neural network (NCC-NN) structure with positive samples, such as different snapshots of the target, and negative samples, such as different types of clutter that create false alarms, a  filter bank is obtained as the weights of a complete layer of the trained neural network. This way, not only the detection success is maximized but the filters that create the minimum false alarm rates are obtained, simultaneously.

Convolutional neural networks have recently become the state-of-the-art de-facto standard of any signal-based machine learning approach. And there are many recent studies that focus on using deep CNNs to detect and recognize various types of objects or targets. The main reason we choose to use the proposed NCC layer, instead of convolutional layers, is that NCC layer needs relatively less data to converge. For many operational concepts, such as detection in search-and-track systems, the amount of available training data is not sufficient to prevent a deep CNN from over-fitting. We discuss the benefits of using the NCC layer instead of the conventional convolutional layer in the following sections. 

We also propose the mean-absolute-deviation NCC (MAD-NCC) layer, an efficient implementation of NCC layer, designed especially for the FPGA systems. In this optimized design, square root operations are avoided for real-time computation and minimal resource usage. As a case study, we work on dim-target detection on mid-wave infrared imagery and benchmark the performance of different filter designs. The results of the trained NCC-NN shows that instead of choosing heuristic filter designs, it is possible to converge to a filter set that would come up with an improved receiver operating characteristics.

The paper is organized as follows: the next section introduces the proposed neural network layers, the NCC layer and its optimized form, the MAD-NCC layer. The dataset used in our experiment is introduced in Section 3. Results are provided in section 4, for a case study on dim-target detection on mid-wave infrared imagery, while the final section outlines the conclusions of this study.

%%%%%%%%%%%%%%%%%%%%%%%%%%%%%%%%%%%%%%%%%%%%%%%%%%%%%%% 

\section{Normalized-Cross-Correlation Layer for NNs}
In this section we introduce the normalized-cross-correlational layer, which is an alternative to convolutional layers in neural networks. In addition, explicit formulation of forward and backward propagation functions for the proposed structure are provided.

\begin{figure}[t]
\centering
\includegraphics*[trim=0 0 0 0,clip=true,width=1\textwidth]{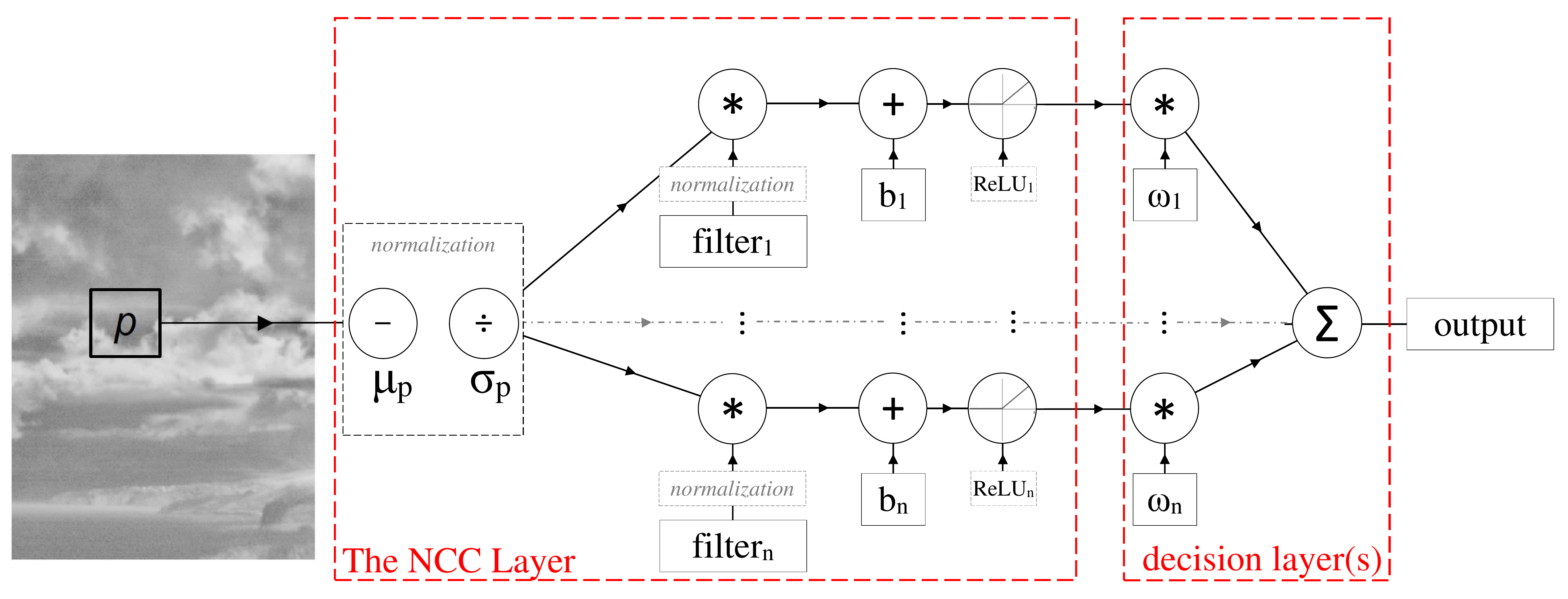}
\caption{A sample two-layer NCC-NN structure, with an NCC layer and a simple fully connected Rectified Linear Unit (ReLU) decision layer, is depicted. NCC layer is similar to a convolutional layer, with the exception that the patch is normalized prior to being convolved with the filters.}
 \label{NCCLayer}
\end{figure}

\subsection{The NCC Layer}
The NCC layer is a variation of a convolutional layer of a neural network, with the exception that the input is normalized prior to being convolved with the filters (or kernels) of that layer. A simple structure of the NCC layer is given in Figure \ref{NCCLayer}. As can be seen from the figure, if the normalization blocks are removed, the NCC layer becomes identical to a convolutional layer. 

The outcome of normalizing the input before convolving it with a filter is simply limiting the output values. When both the filter and the input is normalized, the convolution operation becomes identical to normalized-cross-correlation\footnote{Either the filter or the signal must be real-valued and symmetric for this generalization to hold, which is a general case when CNNs are considered. The pixel values are always real valued and so our filter values. How to constrain the filter to symmetric shape or its (un)necessity for our application of the NCC layer is discussed in the following sections.} and the output is bounded in the range [-1,1]. The output of NCC between two 2D discrete signals \emph{A(i,j)} and \emph{B(i,j)}, defined as $\rho^{2D}_{A,B}$ in Equation \ref{NCC}, is a measure of linear dependence between A and B, with +1 showing complete positive dependence, -1 showing complete negative dependence and 0 showing linear independence. 

\begin{equation}
\mathbf{p} \overline{*} \mathbf{f} = \frac{1}{(n-1)}\cdot \sum_{i}^{ }\frac{\left (\mathbf{p}(i)-\mu_{p} \right )\cdot \left ( \mathbf{f}(i)-\mu_{f}  \right )}{\sigma_{p}\cdot \sigma_{f}}
\label{NCC}
\end{equation}

\noindent
In Equation \ref{NCC}, $\overline{*}$ symbol represents the normalized-cross correlation operation, \emph{n} is the total number of pixels\footnote{Although \emph{p} is a 2D real-valued signal, an image patch, the notation \emph{p(i)} is preferred instead of \emph{p(i,j)} for the sake of simplicity. The variable \emph{i} represents a total of \emph{n} pixels in both horizontal and vertical dimensions of the patch \emph{p}.} in $\mathbf{p}$ or $\mathbf{f}$, \(\mu\)$_{p}$ and \(\mu\)$_{f}$ are the average pixel values and \(\sigma\)$_{p}$ and \(\sigma\)$_{f}$ are the standard deviations of the pixel values of p and f, respectively.

\subsection{Why Normalize?}

One of the main reasons why a normalization is conventionally not preferred for a CNN layer is the fact that bounding the output range may limit, or even diminish propagation. What a Re-Lu layer does in a CNN is \cite{Arora2016} to destroy negative, and proportionally admit positive, forward propagation. Limiting both the negative and positive output of the convolutional node, like it happens for the proposed NCC layer, is similar to using a poor activation function and will result in a poor performance for CNNs. Despite this seemingly undesired fact, we have two main motivations in utilizing the NCC layer. Firstly, the main reason why NCC is a good template matching method for infrared imagery is the fact that, the low signal frequencies (like the average image intensity) are destroyed in normalization. In infrared imagery this is a desired fact, because depending on the dynamic range of an infrared detector, the low frequencies of an infrared image is expected to differ among different systems that uses the same infrared sub-band. Theoretically, a CNN is capable of discriminating these low-band signal properties. However the performance of CNNs, especially deep CNNs, depend on the availability of data. In practical situations, such as military related infrared detection problems, such data are not usually available. In those cases, CNN-based methods may easily overfit, for example, to mean intensity values of a scene (which may depend on the air temperature), when the collected samples are limited. What is more, a detection or recognition capable neural network, trained with a specific dataset, may be (and is most usually) applied to perform for another infrared sub-band or another detector, in which low frequency elements like the mean intensity are expected to differ, as well. Thus, the NCC layer, when infrared detection and recognition tasks are considered, has more generalization power compared to a convolutional layer and is more prone to converge to optimum weights (i.e. filter shape) with relatively limited data.

Secondly, utilizing the NCC layer structure, compatible with a general neural networks architecture is quite beneficial considering the fact that it can easily be trained using back-propagation. This way we can extract the NCC layer as a filter bank and directly utilize it for an operational purpose (detection, recognition, etc.) In order to better experiment the proposed concept, in the following sections we provide a custom NCC layer structure, which can be easily combined with any of the multi-layered deep learning software libraries.

\subsection{NCC Layer Implementation}
An NCC node, as given in Figure \ref{NCCLayer}, is a serial combination of a normalization node and a convolutional node. Equation \ref{NCC} is an explicit formulation of the forward operation of this layer. However, in this chapter we will obtain the forward and backward propagation formulas for this layer by considering the normalization and convolution as two separate sequential operations. The reason we chose to separate these two formulas is practical. Extremely fast GPU-based solutions exist for forward and backward convolution operations in CNNs. Thus, instead of constructing the function for this new layer from scratch, it is practically much more convenient to detach two operations, derive functions for normalization only, append these functions to a convolutional layer of an existing CNN library (such as  MatConvNet \cite{MatConvNet}) and experiment on a desired set of data.  Below in Equation \ref{ForwardNCC}, an equivalent formulation of the NCC forward function is provided.

\begin{equation}
xcorr(p) = \overline{\mathbf{p}} \star \overline{\mathbf{f}}
\label{ForwardNCC}
\end{equation}

The forward function of the NCC is simply the convolution of the normalized 2D discrete signal $\overline{\mathbf{p}}$ with the filter $\overline{\mathbf{f}}$. The normalized $\overline{\mathbf{p}}$ can be calculated as:

\begin{equation}
\overline{\mathbf{p}} {=} \frac{1}{\sqrt{n-1}} \cdot \frac{\mathbf{p}-\mu_{p}}{\sigma_{p}}
\label{Normalization}
\end{equation}

In Equation \ref{Normalization}, $\mu_{p}$ represents the mean pixel value and $\sigma_{p}$ represents the standard deviation of the patch $\mathbf{p}$. Below we also provide the well-known formulas of mean and standard deviation calculation, because we will need to derive their derivatives for backward function calculation in the following. Using \ref{ForwardNCC} and \ref{Normalization}, we obtain the same forward operation that was given in \ref{NCC}, but in an alternative form where normalization is nested within the convolution/correlation operation.

\begin{equation}
\mu_{p} = \sum_{i=1}^{N}{\mathbf{p}(i)} \,\,,\,\, \sigma_{p} = \sqrt{\frac{1}{(n-1)}\sum_{i=1}^{N}{(\mathbf{p}(i)-\mu_{p})^2}}
\label{meanandstd}
\end{equation}

In order to obtain the backward propagation functions of a node in a neural network, we find the partial derivatives of the output with respect to input patch and the filter values (weights) of that node. In our case, the forward function is a normalization function nested within convolution. In order to obtain the back propagation function, we simply apply the chain rule:

\begin{equation}
\frac{\partial O}{\partial \mathbf{p}} = \frac{\partial O}{\partial \overline{\mathbf{p}}} \cdot \frac{\partial \overline{\mathbf{p}}}{\partial \mathbf{p}}
\label{dOdp}
\end{equation}

\begin{equation}
\frac{\partial O}{\partial \mathbf{f}} = \frac{\partial O}{\partial \overline{\mathbf{f_k}}} \cdot \frac{\partial \overline{\mathbf{f_k}}}{\partial \mathbf{f_k}}
\label{dOdf}
\end{equation}

In Equations \ref{dOdp} and \ref{dOdf}, \emph{O} represents the scalar output of an NCC node, $\mathbf{p}$ represents the input patch and $\mathbf{f}_k$ represents the k$^{th}$ filter in that layer. Equations are identical, since both the input and the filter are normalized and the convolution operation is commutative. $\partial O \slash \partial \overline{\mathbf{p}}$ stands for the partial derivation of convolution operation, whereas  $\partial \overline{\mathbf{p}} \slash \partial \mathbf{p}$ stands for the derivative of normalization operation (\ref{Normalization}), which is given below in Equation \ref{devnorm}.

\begin{equation}
\frac{\partial \overline{\mathbf{p}}}{\partial \mathbf{p}} {=} \frac{1}{\sqrt{n-1}\cdot\sigma_{p}} \cdot (\mathbf{I}-\frac{\mathbf{1}}{n}) \cdot (\mathbf{I}-\frac{(\mathbf{p}-\mu_{p})(\mathbf{p}-\mu_{p})^T}{(\mathbf{p}-\mu_{p})^T(\mathbf{p}-\mu_{p})})
\label{devnorm}
\end{equation}

As mentioned above, by using Equations \ref{Normalization} and \ref{devnorm}, we implemented the NCC layer within MatConvNet library \cite{MatConvNet} by appending the forward and backward functions for normalization to a custom layer, which also uses the hardware-based (both forward and backward) convolution functions of this library. By using any other open source deep net library (such as Caffe \cite{Caffe}, etc.), NCC layer can be identically implemented.

\subsection{MAD-NCC Layer Implementation}
The MAD-NCC layer is an efficient implementation of NCC layer, in which mean-absolute-deviation (MAD) operation is used instead of standard deviation operation. Thus, for the MAD-NCC layer implementation,  the input patch $\mathbf{p}$ is normalized by using the equation below:

% tilde for mad-normalization
\begin{equation}
\tilde{\mathbf{p}} {=} \frac{1}{\sqrt{n}} \cdot \frac{\mathbf{p}-\mu_{p}}{mad_{p}}
\label{MadNorm}
\end{equation}

In \ref{MadNorm} , $mad_p$ stands for the mean-absolute-deviation of image patch and is calculated as:

\begin{equation}
mad_{p} = \frac{1}{n}\sum_{i=1}^{N}{|\mathbf{p}(i)-\mu_{p}|}
\label{madfor}
\end{equation}

Consequently, the backward function  $\partial O \slash \partial \tilde{\mathbf{p}}$ for MAD-normalization can be derived as:

\begin{equation}
\frac{\partial \tilde{\mathbf{p}}}{\partial \mathbf{p}} {=} \frac{1}{\sqrt{n}\cdot mad_{p}} \cdot (\mathbf{I}-\frac{\mathbf{1}}{n}) \cdot (\mathbf{I}-\frac{(\mathbf{p}-\mu_{p})\cdot sign(\mathbf{p}-\mu_{p})^T}{n\cdot mad_{p}}
\label{devmadnorm}
\end{equation}

In Equation \ref{devmadnorm}, \emph{sign($\cdot$)} is the signum function. Similarly to the NCC layer, by using \ref{MadNorm} and \ref{devmadnorm}, we implemented the MAD-NCC layer within MatConvNet library \cite{MatConvNet} by appending the mad-normalization (forward and backward) functions into a custom layer and again by using the built-in hardware-based convolution functions of this library.

Our motivation behind designing the MAD-NCC layer is basically to avoid vector-based square-root operations that exist in standard deviation calculation (Equation \ref{meanandstd}). Although there are novel approaches to square root calculation on FPGA \cite{Kachhwal2014}, this operation is relatively slow compared to many other operations, such as division by 2$^n$ (i.e. bit shift). This way, we expect to have a faster forward operation, with a slightly degraded performance. In the following sections we compare the performance of our NCC and MAD-NCC hardware implementations, and discuss the impact of the proposed trade-off.

\section{The Dataset}
The problem of IR target detection is generally specific to target type, target range and camera properties. For instance, in order to detect air targets from long ranges up to 10km, a proper systems engineering study would indicate the need for a camera with a narrow field-of-view (FOV) (smaller than 5$^{\circ}$) and a detector of mid-wave infrared band. Thus, searching for or trying to create a general detection dataset for the entire IR target detection problem would be undesirable and incorrect by definition. What is more, even \emph{small target detection} is a broad definition, which must be narrowed down for proper academic and engineering analysis. 

In order to study the proposed NCC Layer, we have constructed a ``small target database'' (not a small database but a database of small targets) specific to our problem. We used a mid-wave infrared detector and an optics with a narrow FOV\footnote{The details of the optical system and the dataset itself are unfortunately not publicly available, due to the classified nature of the project that belongs to ASELSAN Inc.}. The dataset includes scenes of actual targets that are captured from different ranges so that the targets in pixel coordinates are small (i.e. around a few pixels or sub-pixels). The targets that have been captured are various air platforms, such as UAVs, helicopters, planes, etc. The scenes included high levels of natural clutter such as terrain, sea surface (with glints), clouds and some human constructions (Please check Figures \ref{DatasetFigure}). 

In addition to real-world scenes, images of point targets captured by using the actual optical system and a collimator \cite{Shafer93} are also included in the dataset. This system provides perfect scenes with virtually no clutter. However, these environments provide simple but undisturbed target signals with the actual detector characteristics, such as the detector noise and bad-pixels. IR detectors provide pixels with non-uniform transfer functions and for this purpose require a non-uniformity correction (NUC). Depending on the NUC method, some pixels may be selected as "bad" (i.e. having bad transfer functions) or dead (i.e. unresponsive) and can be replaced with the adjacent pixels \cite{Mudau2011}. However, even the best NUC and bad pixel replacement (BPR) methods leave undetected bad pixels and in most cases these bad pixels are very similar to real small targets. Thus, including collimator scenes in our dataset helps us to study these detector characteristics as well.

\subsection{Dataset Preparation}
\begin{figure*}[t]
\centering
\begin{subfigure}{4.8cm}	
	\includegraphics*[width=4.2cm, height=5.6cm]{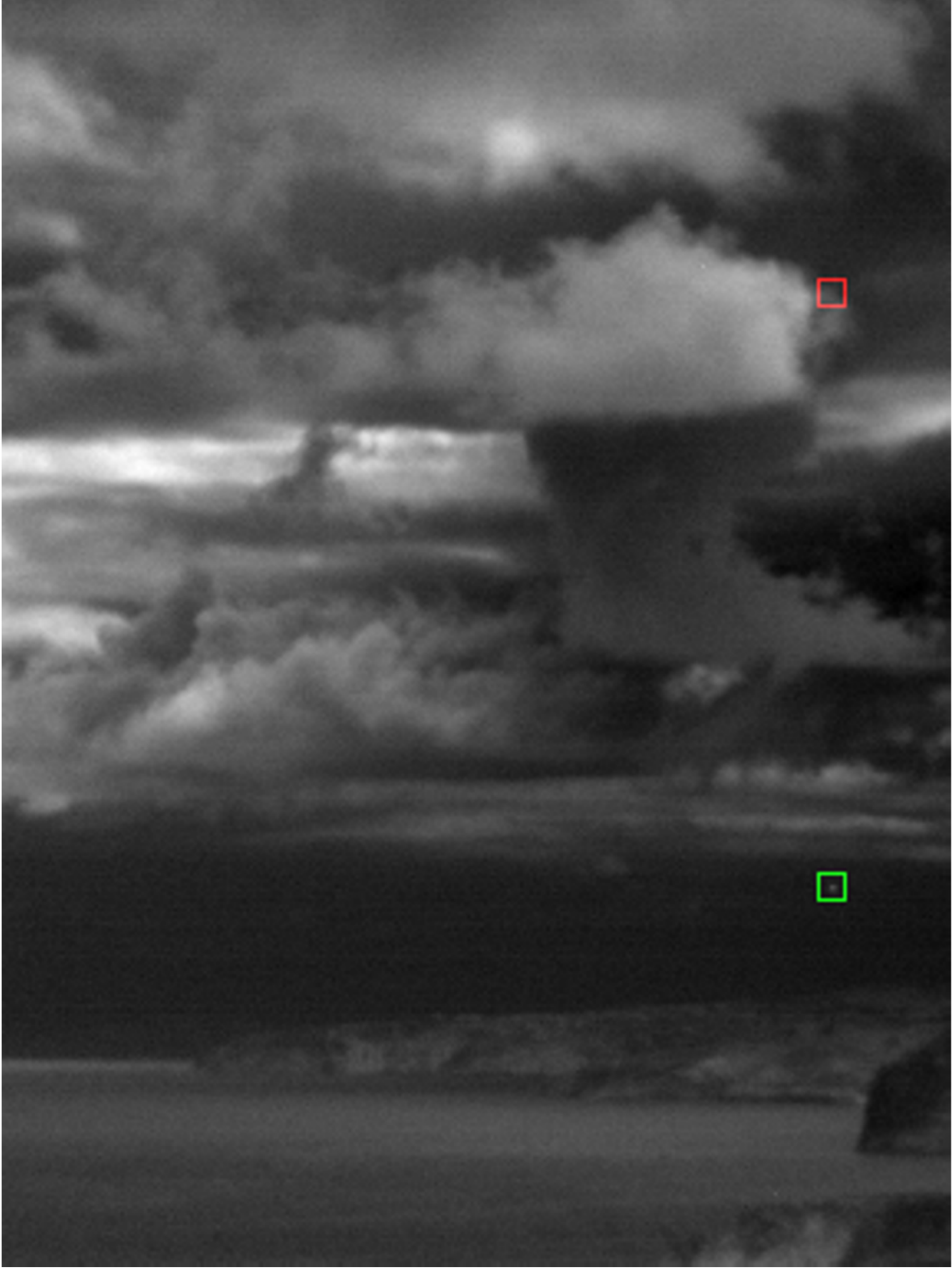}
	\caption{A sample dataset Image}	
\end{subfigure}
\begin{subfigure}{4.0cm}	
	\includegraphics*[width=0.95\textwidth]{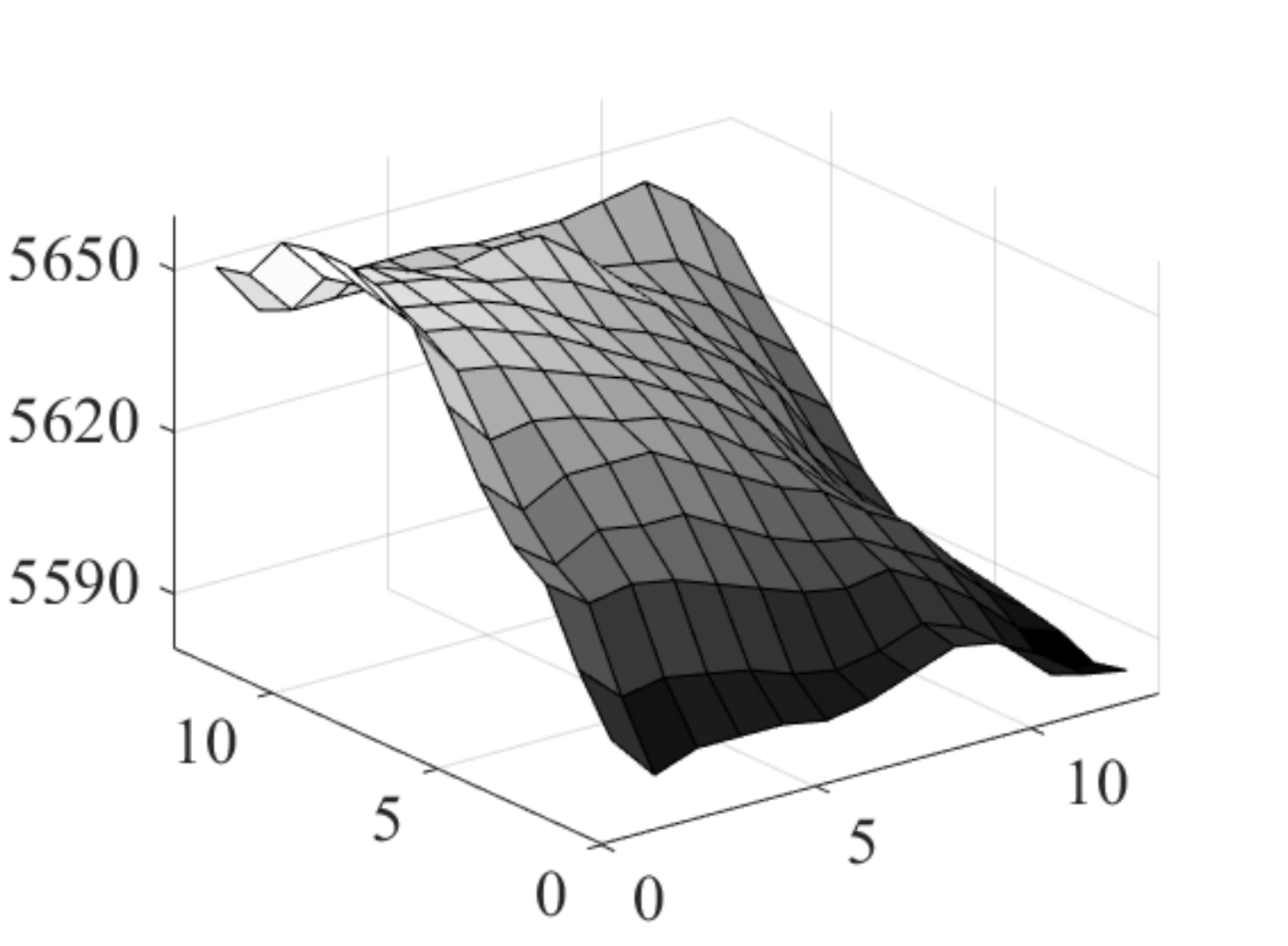}
	\includegraphics*[width=0.95\textwidth]{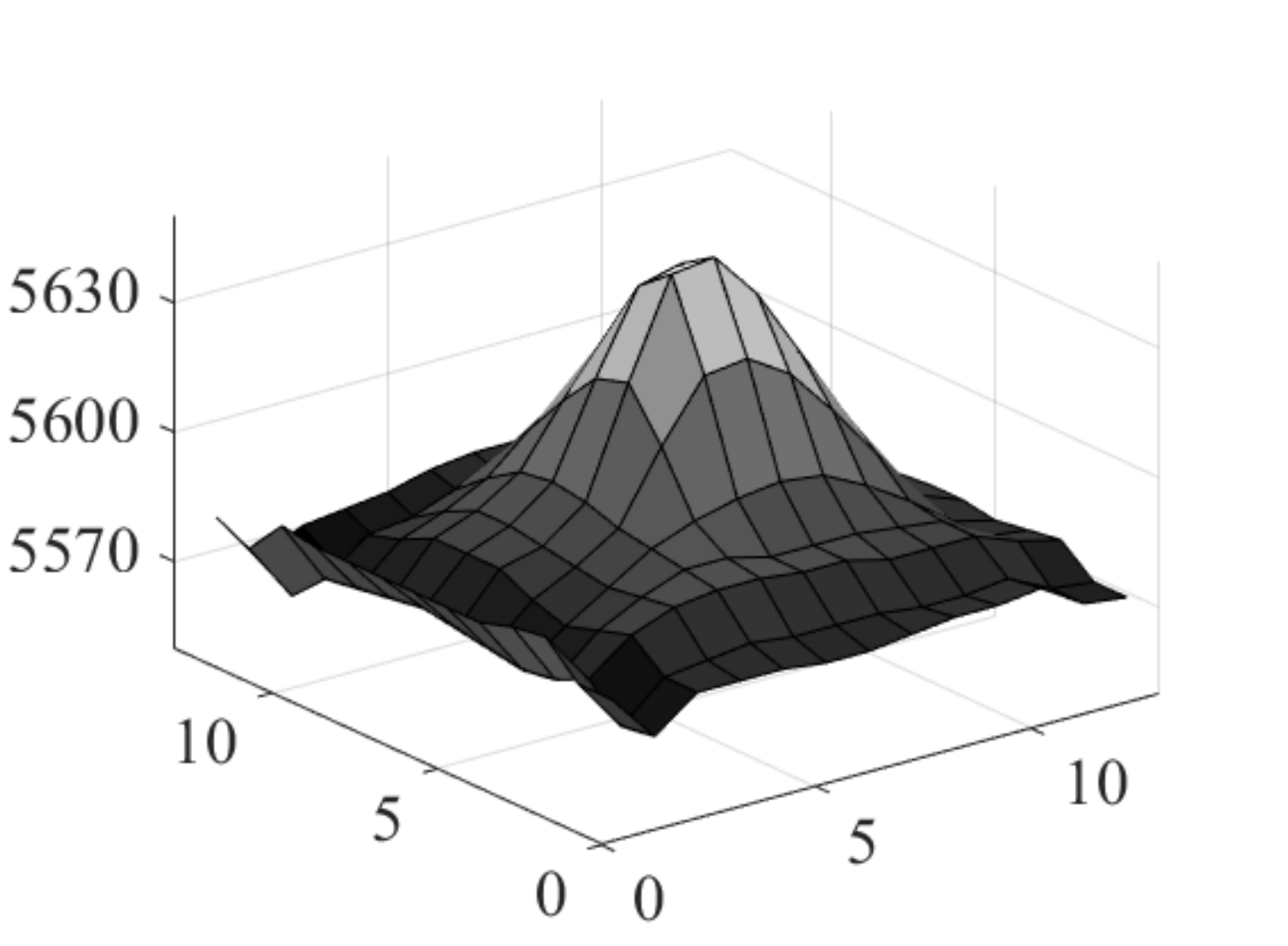}
	\caption{Surface Plots}	
\end{subfigure}
\begin{subfigure}{3.2cm}	
	\includegraphics*[width=2.8cm, height=5.6cm]{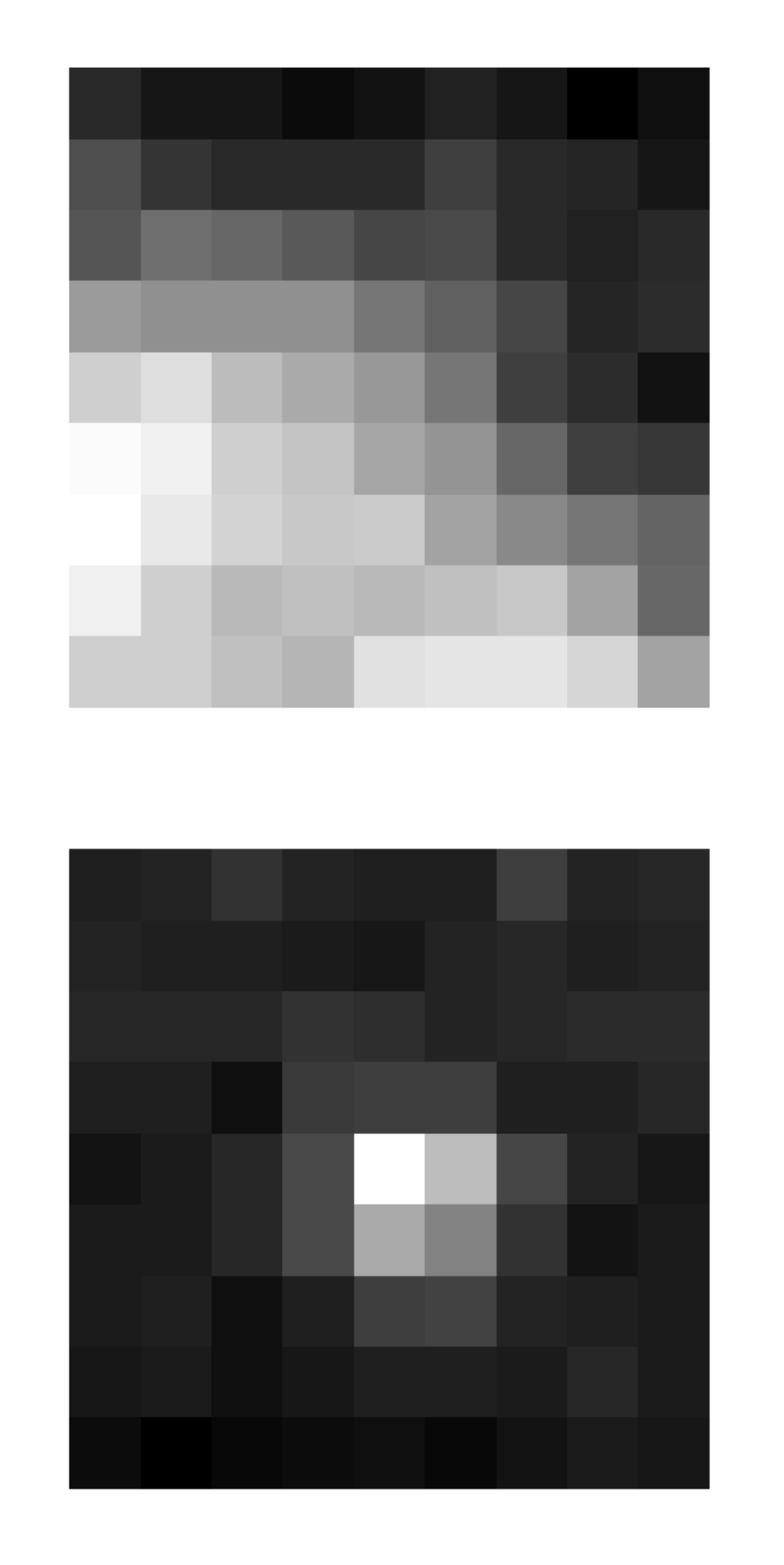}
	\caption{Image patches}	
\end{subfigure}
\begin{subfigure}{3.4cm}	
	\includegraphics*[width=2.8cm, height=5.6cm]{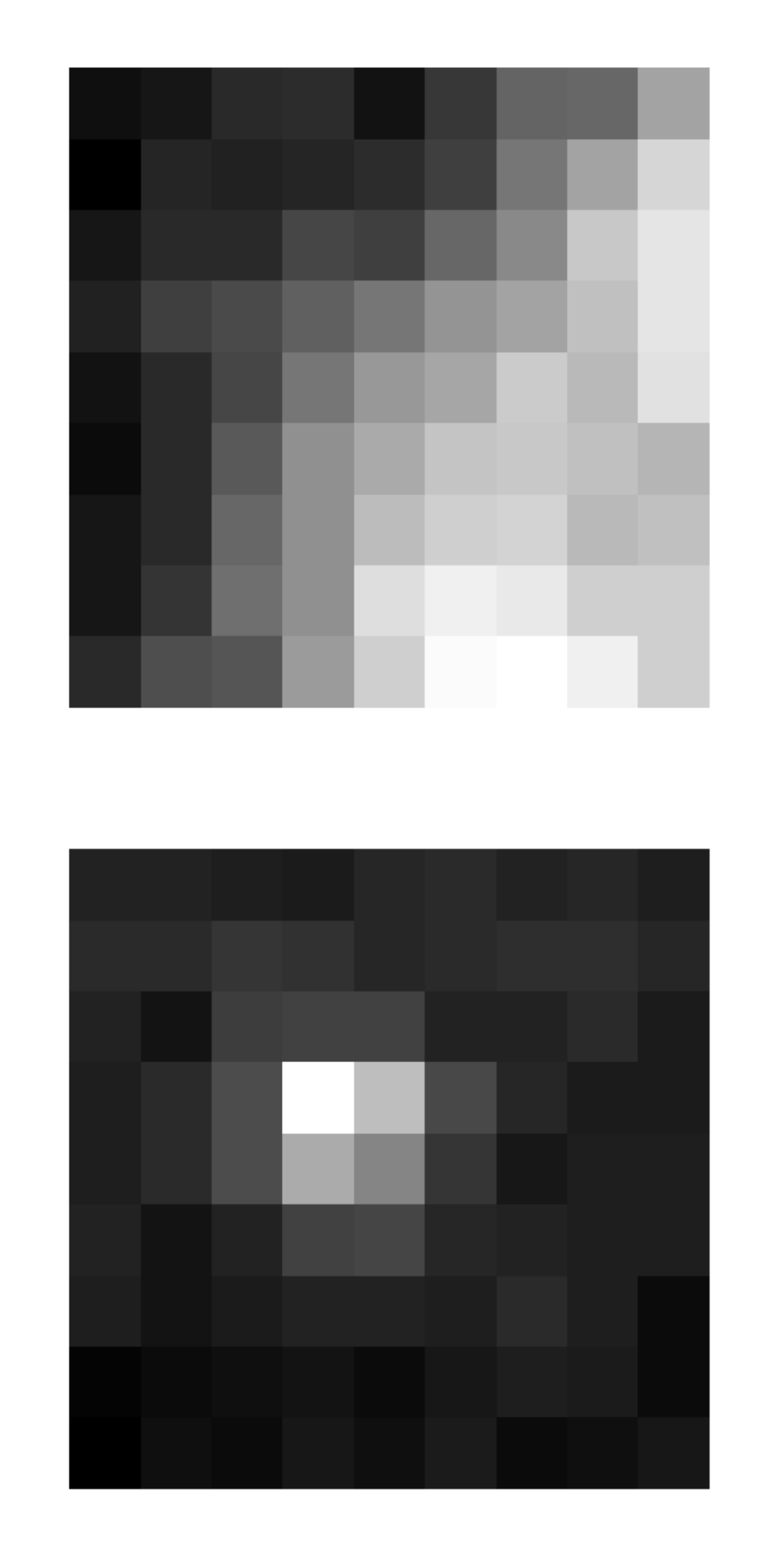}
	\caption{Augmented Patches}	
\end{subfigure}
\caption{In (a), a sample dataset image with a sample 15$ x $15 background region (the red square) and a 15$ x $15 dim-target region (the green square) are depicted. The surface plots for the background (up) and the target (down) are shown in (b). The background (up) and target patches (down) are shown in (c). In (d), augmented versions (up: a rotated sample for the background patch, down: a shifted version for the target patch) are given. }
 \label{DatasetFigure}
\end{figure*}

The dataset includes 4843 frames from 7 different scenes, 2 of which belong to collimator scenes and the rest captured as  real world scenes (Please see Figure \ref{DatasetFigure}a). However, the dataset that we utilize, or in other words, that we feed into our proposed learning structure is not the collection of these image frames. Instead, we extract non-overlapping 15$ x $15 image patches (Figure \ref{DatasetFigure}b) from these approximately 4843 images and use them as the actual input of the proposed learning framework. We manually annotate the patches with actual targets as positive samples. Thus, we end up with 5047 15$ x $15 patches including various small targets (some collimator scenes included multiple targets, thus it is more than 4843). These target patches are selected such that the small target is located at the center of the 15$ x $15 patch (Please see Figure \ref{DatasetFigure}c). 

The rest of the patches, which do not include any small targets, are recorded as the background patches. More than two million background patches were obtained. Having around 5000 positive samples and 2 million negative samples is an extreme case of an unbalanced dataset. Thus, we have applied a very basic correlation-based distance analysis to intelligently select a subset of the background patches, which included all different types of background clutter we come across in our images, such as sky, clouds, terrain, sea surface, collimator noise, bad-pixels, etc. This way we have shrunk our negative samples set to 123385 number of 15$ x $15 patches.

\subsection{Data Augmentation}
The data that is fed into a convolutional neural network is usually raw. In other words, no feature extraction is carried out, so that the network trains itself to find and utilize the necessary features to accomplish the classification task. Very similarly, in our small target detection problem, the proposed NCC-based network accepts raw image patches (i.e. pixel values) to detect targets. In these kinds of networks, the framework do not inherently have any kind of invariance property, such as transformation invariance, illumination invariance, etc. In these cases, data augmentation is utilized to train the system in a desired invariant manner.

For this purpose, we have created the shifted and rotated versions of positive and negative samples of the 15$ x $15 patch dataset. For each positive sample (i.e. patches with a small target), we have added 3 rotated versions (by rotating 90$^{\circ}$, 180$^{\circ}$ and 270$^{\circ}$). For each rotated version (and the original patch), we created 1 pixel and 2 pixels shifted versions of the patches. A shifted positive sample can be seen in Figure \ref{DatasetFigure}d. Thus for each patch, using the rotated (4 orientations) and shifted (16 shifts obtained by a combination of $\pm$1,$\pm$2 pixel shifts in both x and y axes) combinations of the patches, the positive sample set is enlarged by 64 times, creating approximately more than ~326.000 patches with a small target. Augmenting shifted and rotated versions of positive samples is crucial for our method, because the correlation filter is inherently not invariant to shifts and rotations. 

For the negative samples, only rotation is implemented for the purpose of augmentation. Thus, approximately half million negative samples were obtained. Examples of an original and rotated background patches can be seen in Figures \ref{DatasetFigure}c and \ref{DatasetFigure}d.

\section{Experimental Work}
In this section, we provide the details of the experimental work and we commence by introducing the NCC structure used in our experiments. 

\subsection{Two-Layered NCC Network}
The main motivation of this study actually comes out of a practical need to answer the question: "\emph{What is the minimum number and size of correlation filters that is required to discriminate a small target from any type of clutter we come across?}". The number of filters that can be used in our FPGA-based system is limited and thus finding the shape(s) of the filter(s) to accomplish this task is our main interest for this case study.

For this purpose, we have designed a simple two-layered structure using the proposed NCC layer. The first layer, as depicted in Figure \ref{NCCLayer}, is the NCC layer with \emph{at most} four filters. The maximum number of filters is chosen relatively small when compared to state-of-the-art CNN structures. The reason behind this limitation is mainly because of the fact that for the FPGA-based small target detection system we want to develop, we simply cannot afford to have more than a total of four 15$ x $15 correlation filters that can work real-time and together in our system. We wish to observe if there is a feasible solution to the problem with these requirements and hopefully obtain satisfactory results with a less number of filters. Following the first layer, the results of each correlation is fed to a \underline{separate} rectified linear unit (ReLU), so that no negative correlation results is propagated in forward direction and a nonlinear discriminative function can be obtained. The second layer is a single decision layer which is just a weighted sum of the ReLU outputs, again as depicted in Figure \ref{NCCLayer}.  A simple representation of the proposed NCC Network is given in Table \ref{NCCNet}.

\begin{table}[t]
\centering
\begin{tabular}{|l|c|r|}
\hline
\textbf{Input Layer} & \emph{IR patch} & 15$ x $15\\ \hline
\textbf{NCC Layer} & (N$\leqslant$4) \emph{ filters} & 15$ x $15$ x $N\\ \hline
\textbf{Decision Layer} & \emph{weights} & 1$ x $1$ x $N \\ \hline
\textbf{Output Layer} & \emph{targetness measure} & scalar \\ \hline
\end{tabular}
\caption{The Two-Layered NCC Network designed for small target detection.}
\label{NCCNet}
\end{table}

As it can been seen in Table \ref{NCCNet}, both the input patches and the NCC filters have 15$ x $15 pixels size. Moreover, since there is no padding implemented for the NCC-layer, the decision layer has the size of 1$ x $1$ x $N, where N$\leqslant$4 is the number of filter outputs. These layers are implemented as custom layers in MatConvNet library \cite{MatConvNet}. Apart from the convolution function (and its derivate), which is both CPU and GPU compatible in MatConvNet, all other forward and backward operations are implemented as explained in Section 2.3.

\subsection{Training}
The output of the decision layer is a single real value representing the confidence value for having a target within the input patch. Training this structure corresponds to a binary classification problem. To solve this problem, we assign +1 value for any training patch (positive sample) with small target and -1 value to any training patch (negative sample) with no small targets but clutter or noise. Then using  l$_1$-norm as the loss function, the network is trained by back-propagation. Batch normalization is used with a batch size of 40 patches. The training\footnote{{Stochastic Gradient Descent (SGD) algorithm with momentum is employed, considering Momentum: 0.95, Initial Learning rate: 0.001 , Weight Decay: 0.0005.}} is performed, using MatConvNet library \cite{MatConvNet}, on a desktop system with dual 2.6Ghz processors and GPU support of 2880 cores.

In order to train the proposed network, 80\% of the patches are used for each experiment. Thus, for each experiment approximately 260.000 positive samples and 400.000 negative samples are used. Each experiment is executed five times, using a different image subset containing 80\% of the whole dataset.

\begin{figure*}[t]
\centering
\begin{subfigure}{3.2cm}	
    \centering
	\includegraphics*[width=0.9418\textwidth]{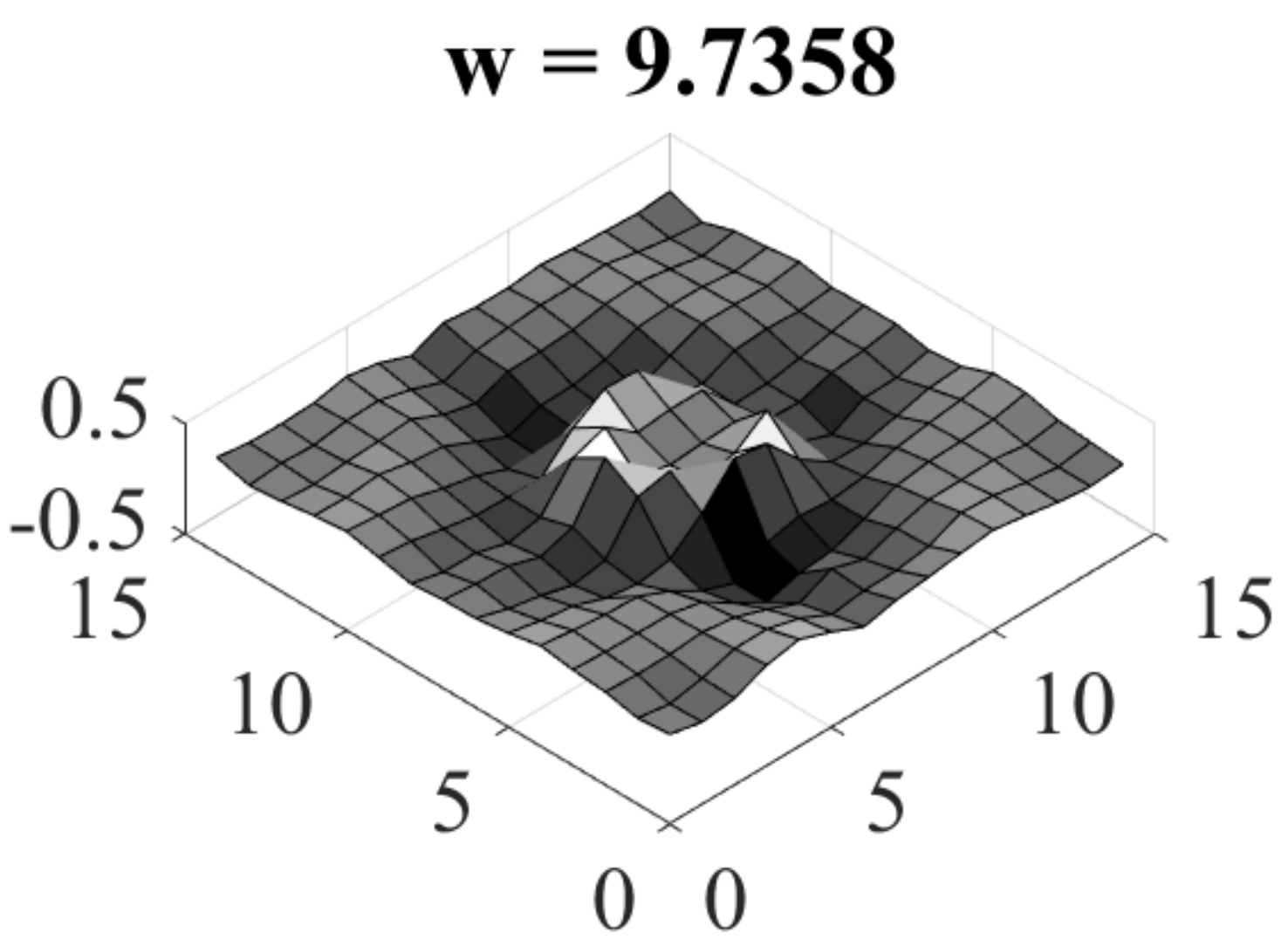} 
	\includegraphics*[width=0.62\textwidth]{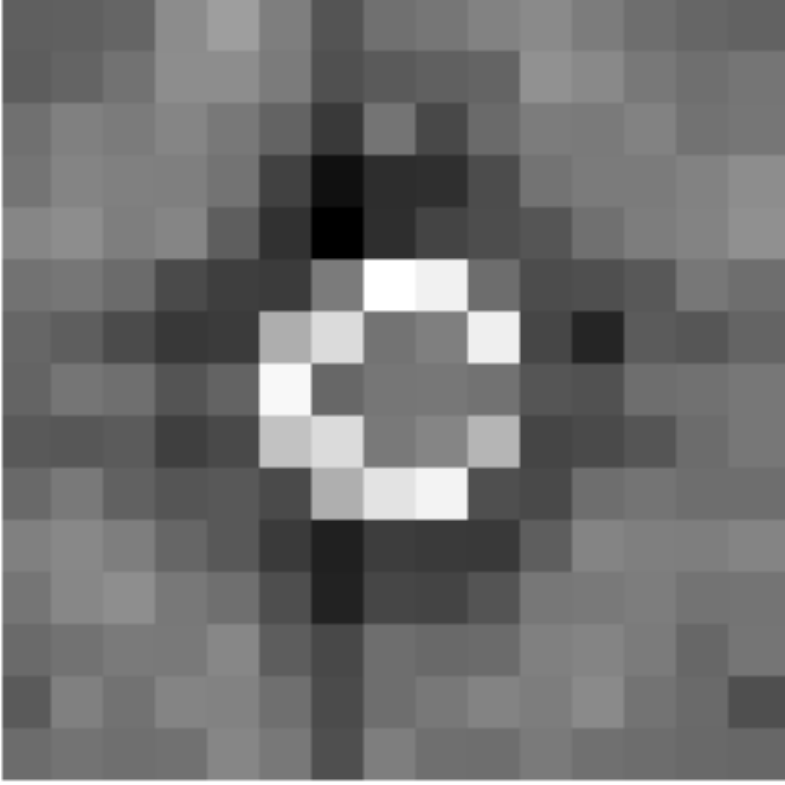}
	\caption{}	
\end{subfigure}
\begin{subfigure}{3.2cm}	
     \centering
	\includegraphics*[width=1\textwidth]{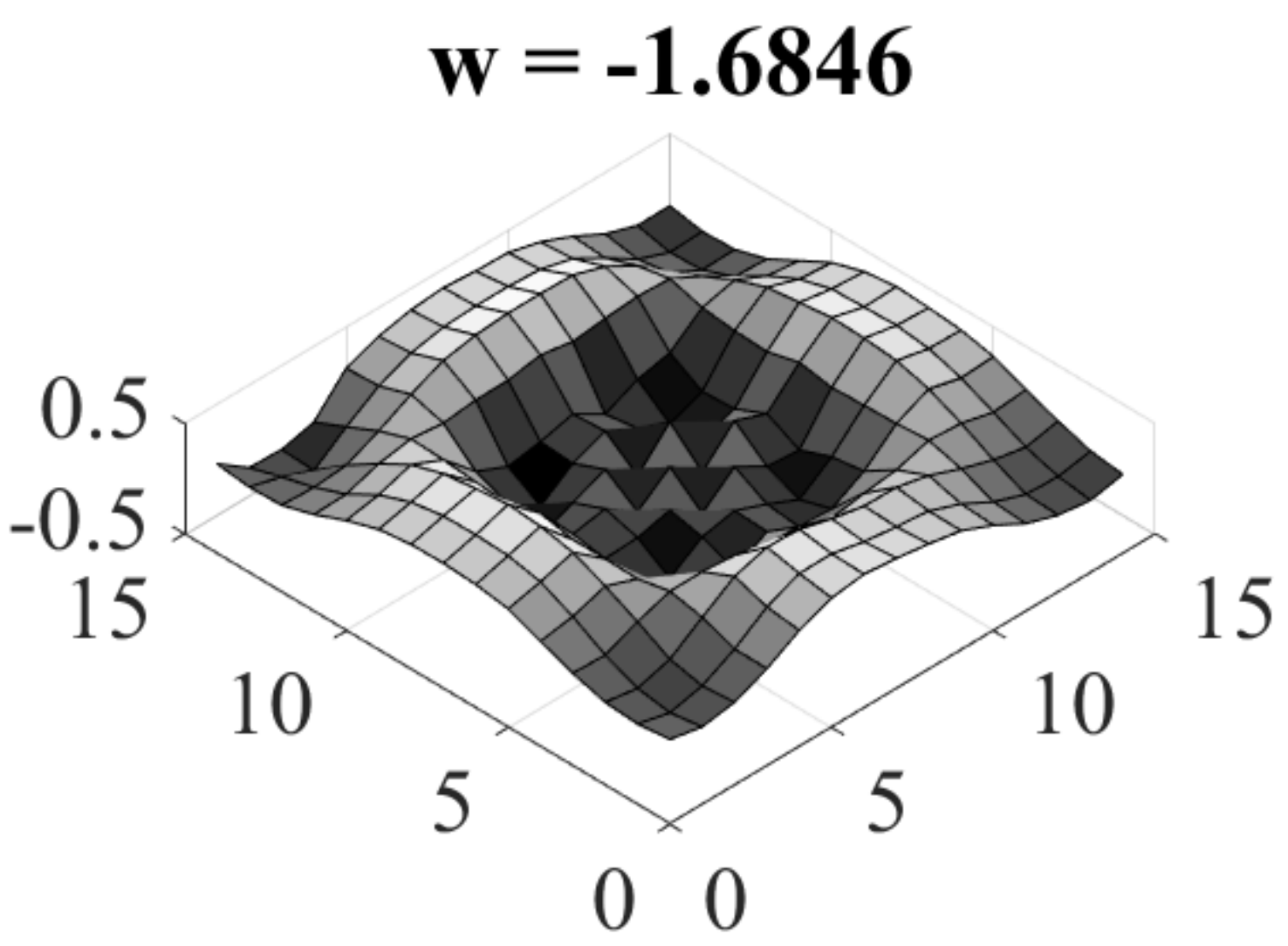}
	\includegraphics*[width=0.65\textwidth]{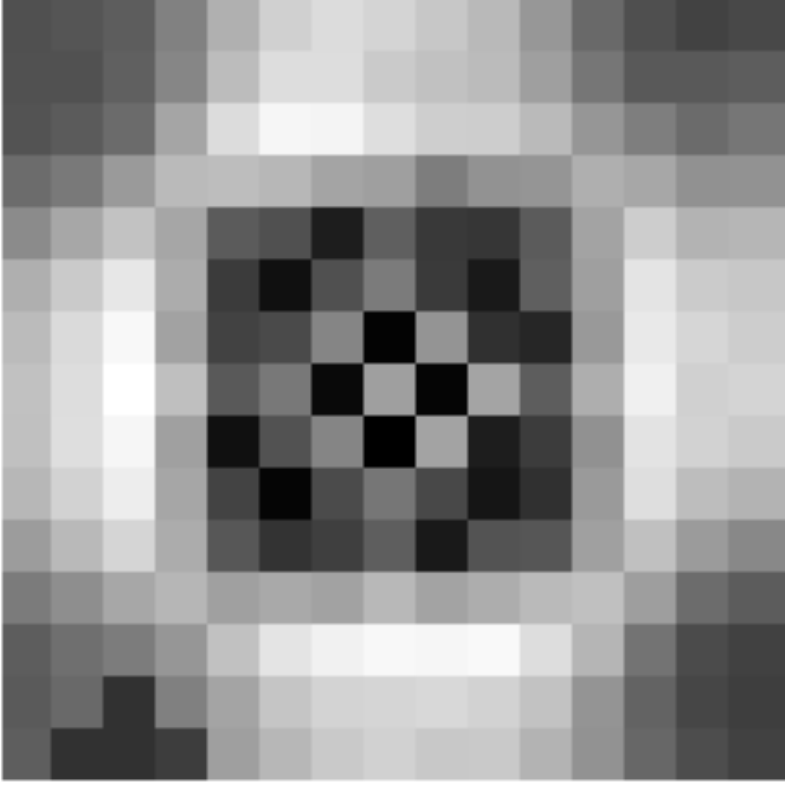}
\caption{}	
\end{subfigure}
\begin{subfigure}{3.2cm}
    \centering
	\includegraphics*[width=1\textwidth]{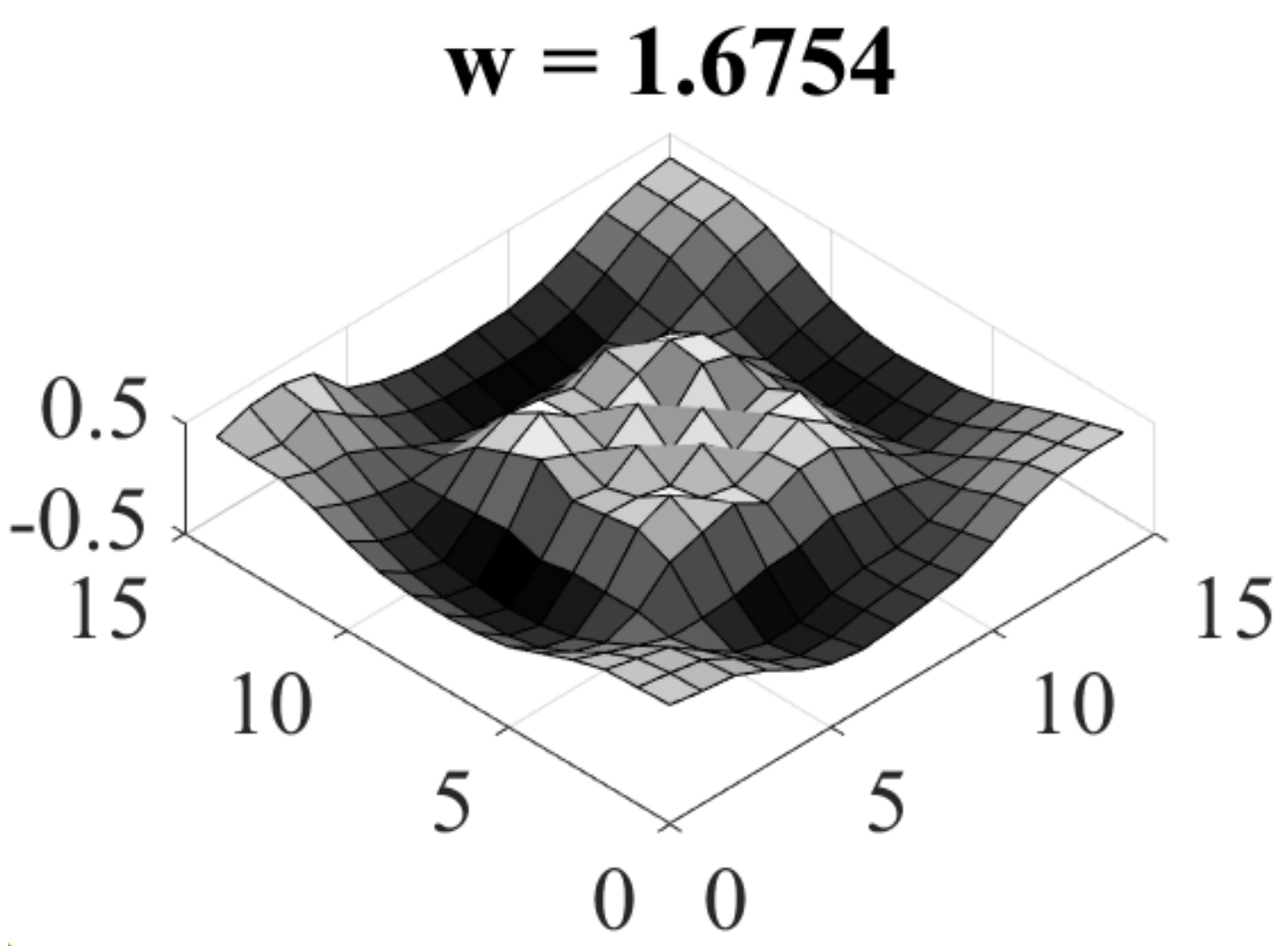}
	\includegraphics*[width=0.65\textwidth]{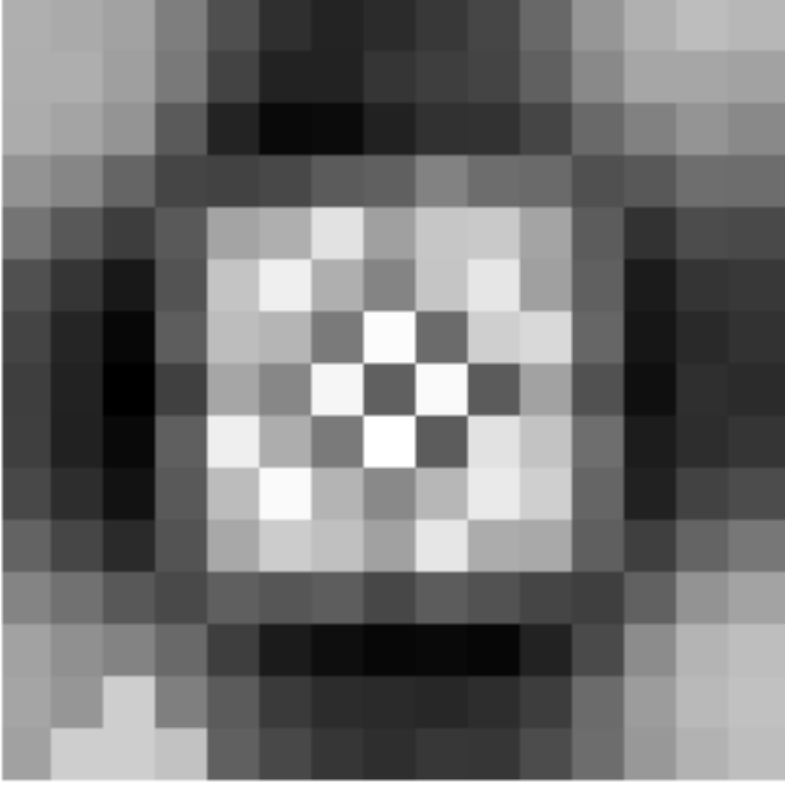}
\caption{}	
\end{subfigure}
\begin{subfigure}{3.2cm}
     \centering
	\includegraphics*[width=1\textwidth]{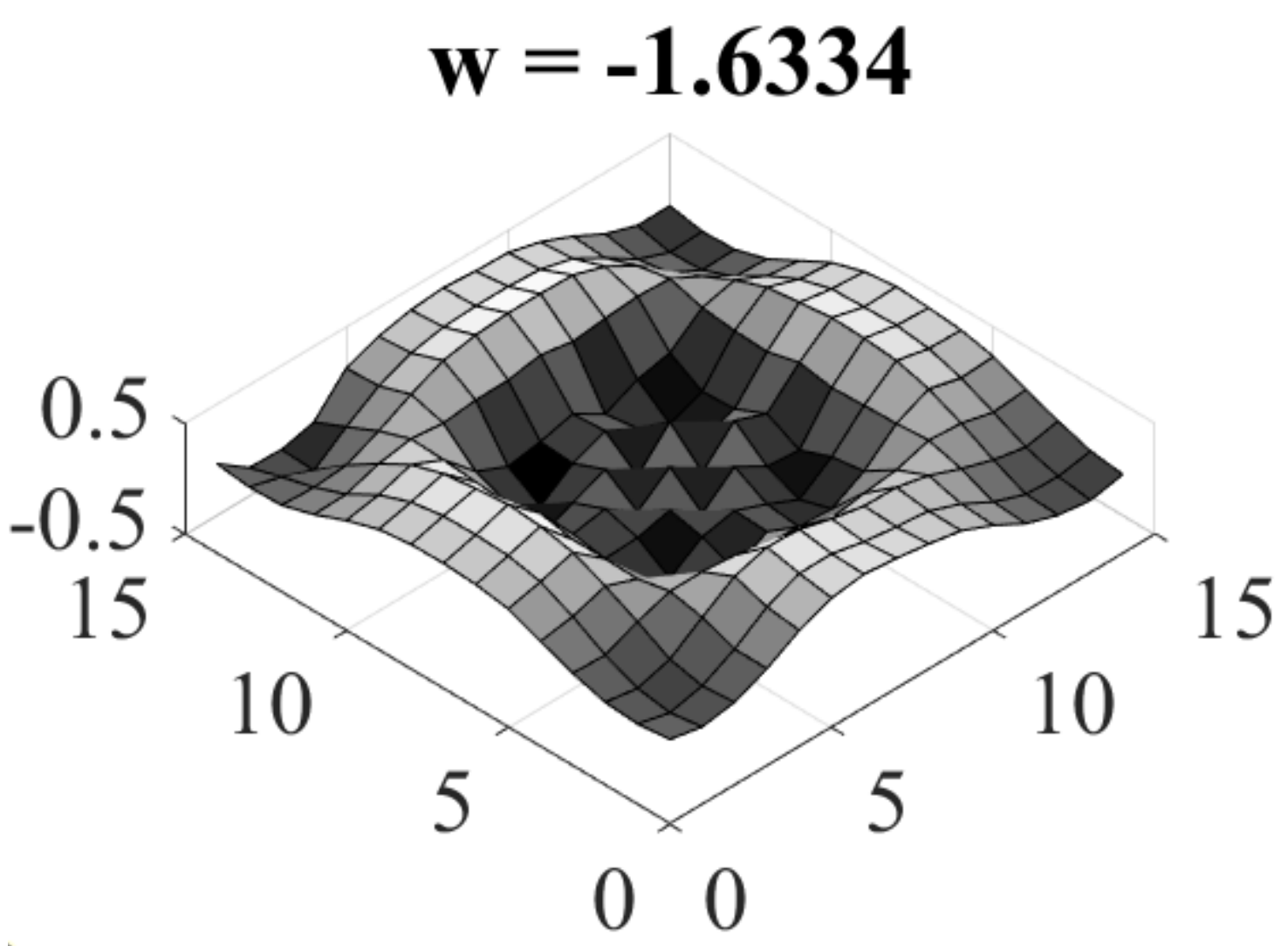}
	\includegraphics*[width=0.65\textwidth]{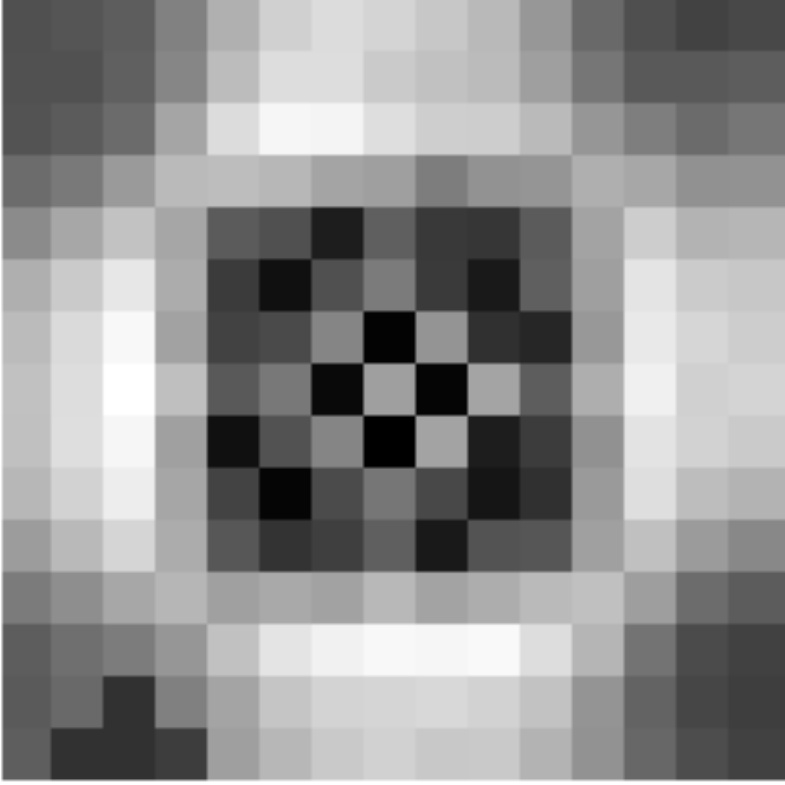}
\caption{}	
\end{subfigure}
\begin{subfigure}{3.2cm}
     \centering
	\includegraphics*[width=1\textwidth]{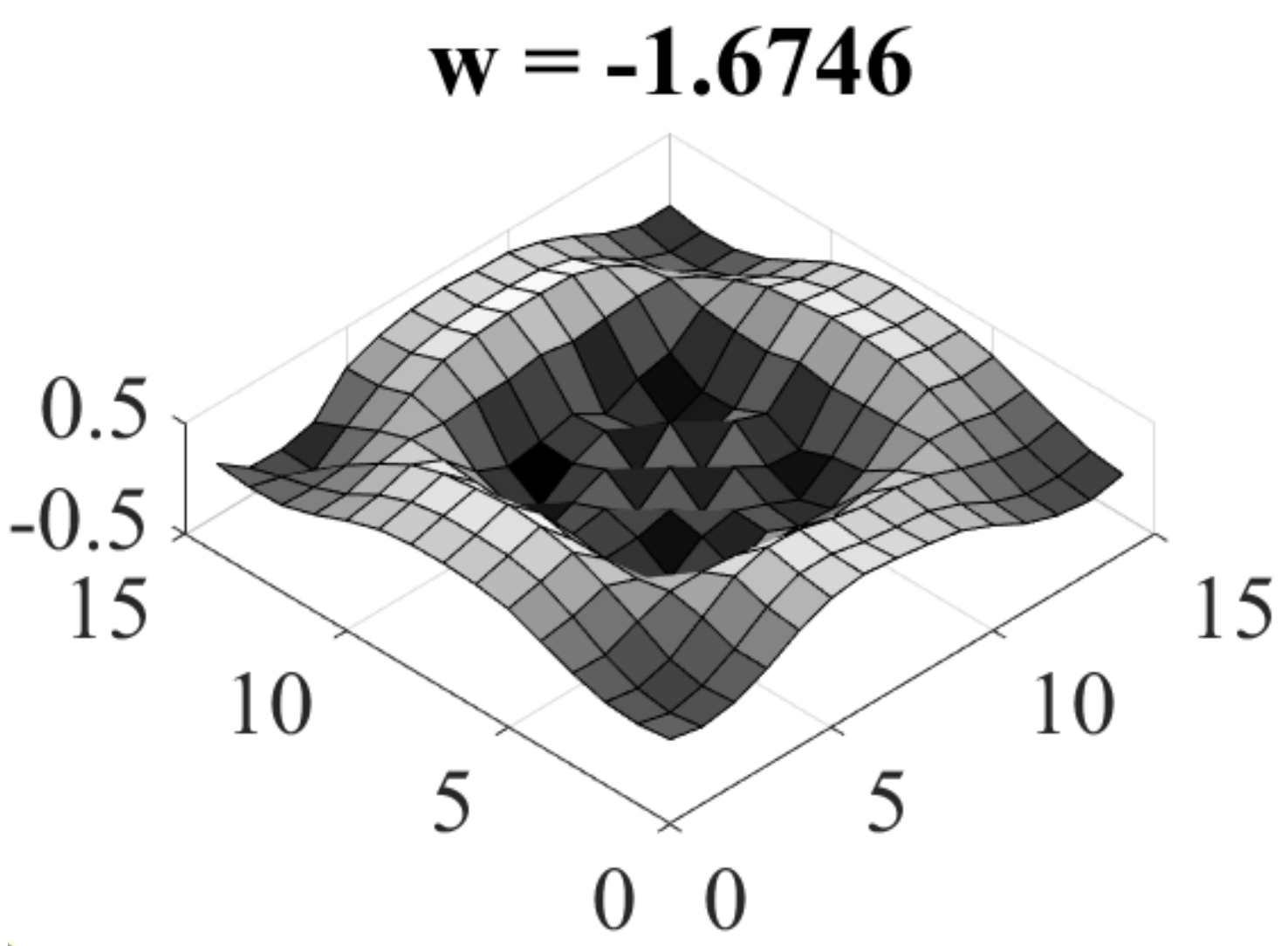}
	\includegraphics*[width=0.65\textwidth]{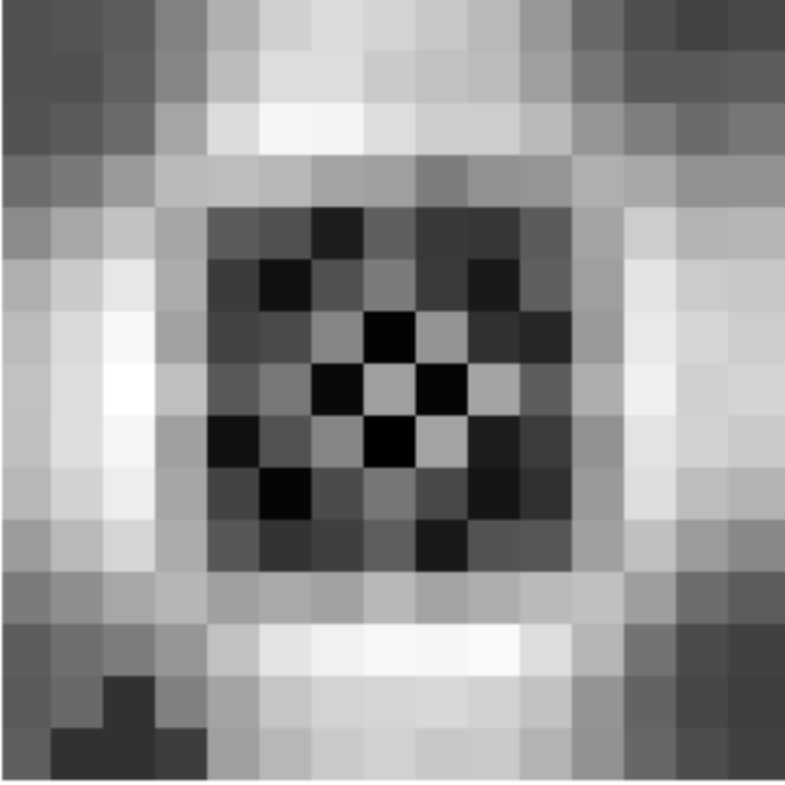}
\caption{}	
\end{subfigure}
\caption{In (a) the resulting 15$ x $15 filter for an experiment with N=1 and in (b,c,d,e) the resulting filters for an experiment with N=4 are shown, as 3D surface plots (above) and gray value 2D images (below).}
 \label{FiltersFigure}
\end{figure*}

\subsection{Resulting Filters}
For different values of N (from 1 to 4), the proposed structure is repeatedly trained. Regardless of the number of filters, in all experiments the system converged to (a) stable filter(s) before 5 epochs. In Figure \ref{FiltersFigure}a the resulting 15$ x $15 filter for an experiment with N=1 and in Figures \ref{FiltersFigure}b, \ref{FiltersFigure}c, \ref{FiltersFigure}d and \ref{FiltersFigure}e, the resulting filters for an experiment with N=4 are shown.

The experiments with a single filter always resulted in a shape very similar to what we see in Figure \ref{FiltersFigure}a. This shape looks very much like a Mexican hat filter, with one exception that at the center of the filter there is a small pit.

The results of an experiment with four filters can be seen in Figures \ref{FiltersFigure}b, \ref{FiltersFigure}c, \ref{FiltersFigure}d and \ref{FiltersFigure}e. Similarly to the case when N=1, the filters look very much like the Mexican hat filter with a pit at the center (or the negated version of this shape). In addition, for almost all runs when N=4, some of the resulting filters were almost identical to each other, such as the filters in Figures \ref{FiltersFigure}b, \ref{FiltersFigure}c, and \ref{FiltersFigure}e., which leads us to think that using 4 filters is unnecessary and redundant. 

When we considered all experiments, regardless of the number of filters, we always obtained a version of this Mexican Hat-looking shape, with a pit at its center. Heuristically, this shape makes a lot of sense. The center is similar to a Gaussian with a pit. The Gaussian shape is to detect small-targets, whereas the center pit is to eliminate single pixel deformations, such as bad pixels. On the other had, the perimeter of the shape, a zero-sum wave is to eliminate clutter, which is usually not uniform. 

These results show that this Mexican-Hat like shape is the filter shape that we are looking for. Moreover, the results show that a single filter is sufficient in discriminating the clutter from the small targets in our dataset, because even with larger N, we always had a version of this filter in the results. In this study we will call this shape "\emph{The Hat}" filter. 

\begin{figure*}[t]
\centering
\begin{subfigure}{8.2cm}	
\centering
	\includegraphics*[width=0.6\textwidth]{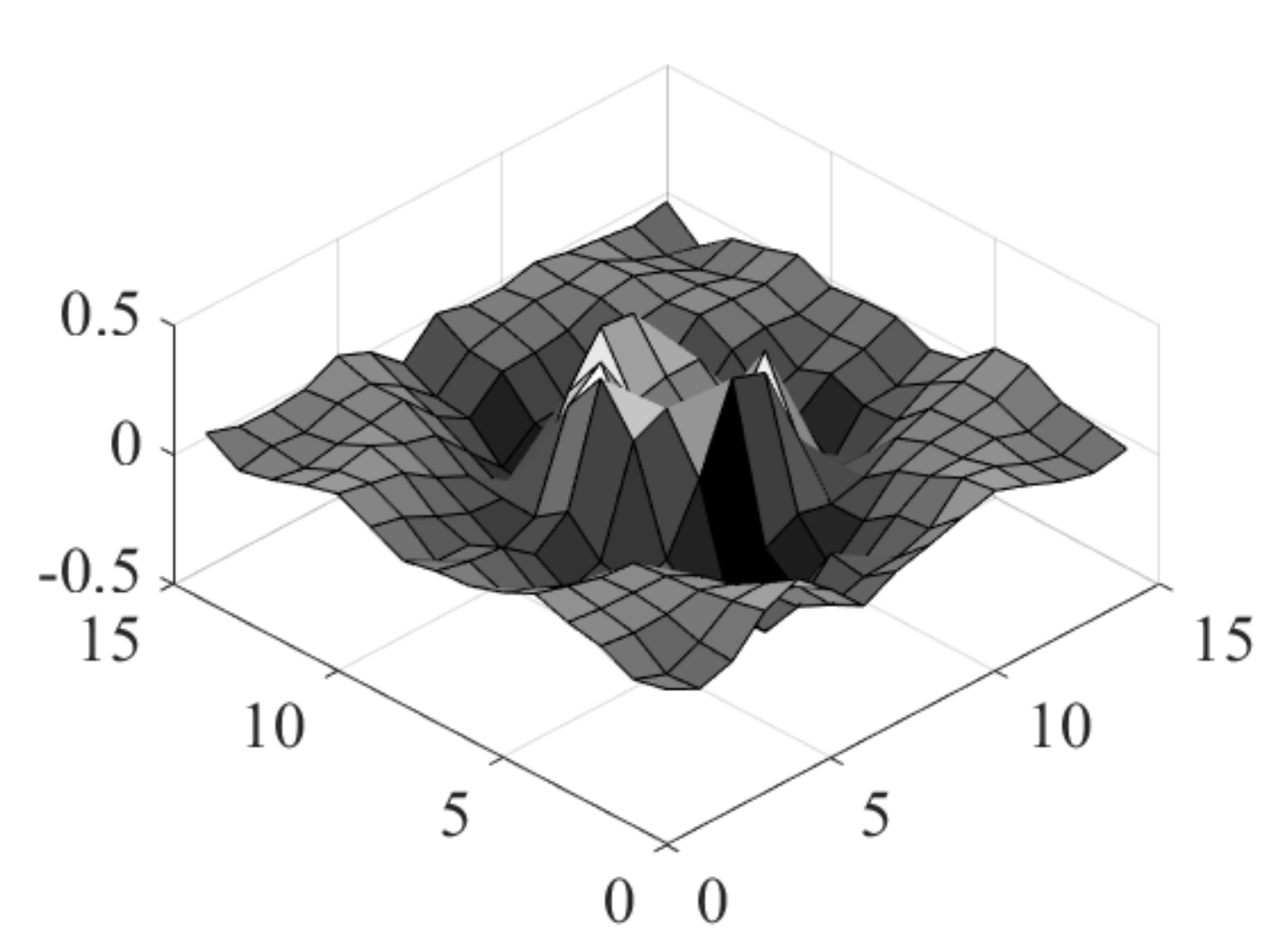}
	\includegraphics*[width=0.38\textwidth]{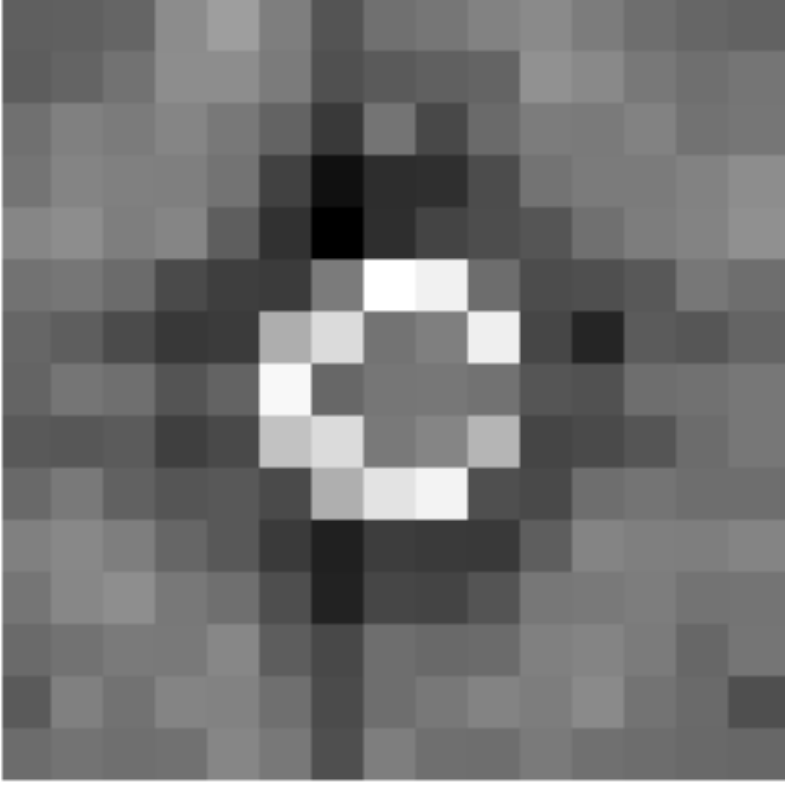}
\caption{}	
\end{subfigure}
\begin{subfigure}{8.2cm}	
\centering
	\includegraphics*[width=0.6\textwidth]{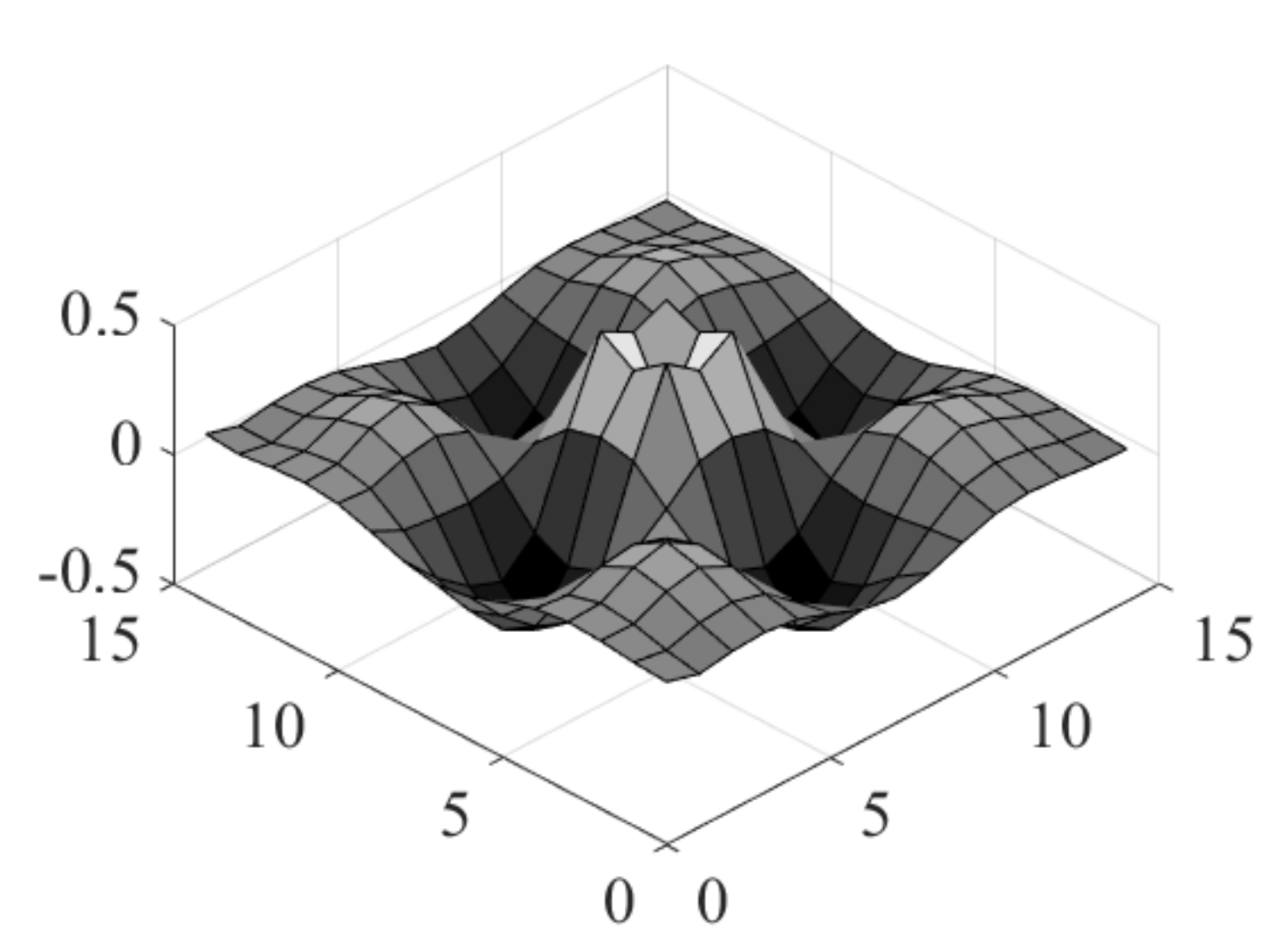}
	\includegraphics*[width=0.38\textwidth]{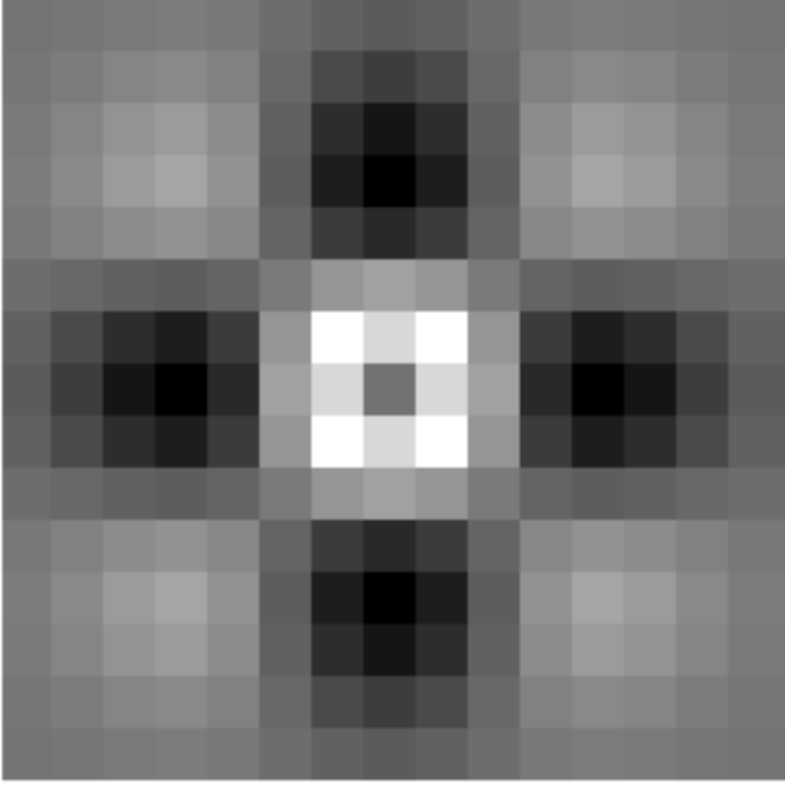}
\caption{}	
\end{subfigure}
 \caption{In (a) The result of one of the N=1 experiments can be seen both 3D (left) and 2D (right) plots. In (b) the so-called "The Hat" filter is depicted.}	
 \label{FilterSums}
\end{figure*}

Instead of using the output of the any of the experiments (such as provided in Figure \ref{FilterSums}a), we generated \emph{The Hat} filter automatically by modifying a Mexican Hat filter in way that it is most similar to the results of the N=1 experiments we obtained (i.e. with a simple optimization routine which would determine the interval of the Mexican Hat wavelet in MATLAB). The so-called \emph{The Hat} filter can be seen in Figure \ref{FilterSums}b. The reason we have decided to generate a filter, instead of directly using one of the experiment results, is to avoid over-fitting and to guarantee rotation invariance. The generated shape has perfect radial symmetry (imposed by the optimization procedure) and thus invariant to rotations. 

\subsection{Benchmarking}
The experimental results show that the so-called \emph{The Ha}t filter is sufficient to discriminate a small target from various type of clutter and noise. In order to further analyze the strength of  \emph{The Hat} filter compared to other filters and to optimize its implementation on an FPGA platform, in this section we provide the benchmarking of different single detection filters. The listed filters below are tested for the entire dataset and receiver-operating characteristics are presented in the following section. Please refer to Table \ref{complexity} for algorithmic complexities of the benchmarked methods. 
\begin{enumerate}
\item \emph{MAD-Ratio}: This method simply checks the deviation, or mathematically speaking the absolute difference, of each pixel from the mean of the patch that the pixels belongs to. For each pixel, the ratio of its individual difference from the patch mean value, to patch mean-absolute-deviation, is calculated. If the ratio is larger than a threshold (2.5 in our experiments) the pixel is chosen to be a (part of a) target. This method is chosen because it is more FGPA friendly than any correlation filter-based method.
\item \emph{Gaussian Filter}: Checking the correlation of a Gaussian filter is the most well-known method used \cite{Warren2002,Barnett1989,Nitzberg1979,Schmidt1990,Wen2003} in small target detection. Although the Gaussian shape assumption for a small target is highly heuristic, it is realistic considering that the target signal is a convolution of many independent transfer functions such as atmosphere, optics, detector response, etc. For this reason, we have included a group of Gaussians with different standard deviations (0.5, 1.2 ans 2.0 pixels in our experiments) in our benchmarking.  
\item \emph{The Hat Filter}: This is the main result that we obtained from experiments with the NCC Network in Section 4.3. For repeated experiments trained with different subsets of the dataset, the filters always converged to the so-called \emph{The Hat} filter. This result is mostly reasonable since the mid part of \emph{The Hat} is a Gaussian and it detects the small target, whereas the side-waves are for eliminating negative patches with background clutter. The pit at the center, we presume is the result of the bad-pixels in our dataset, which are negative samples. 

In order to further optimize this filter for the FPGA platform we have created different versions by considering the following:
\begin{enumerate}
\item \emph{Filter Size N$ x $N}: This is the cropped filter size. For example 9$ x $9 filter refers to the cropped version of the original 15$ x $15 The Hat filter, where 3 pixels are trimmed from four sides. In other words this is not down-sampling, but trimming. The reader should note that the size of the filter is very crucial for FPGA implementation.
\item \emph{Precision}: We also simulated fixed-point calculations in MATLAB, in order to observe the real results of the FPGA platform. So in this section, if the filter is referred to as "fixed", it simply shows that fixed points calculations are used in the simulation. Otherwise, it is referred to as "ideal", which indicates MATLAB's double precision. 
\item \emph{STD vs MAD}: To calculate the normalized cross correlation between two patches, we need to find the standard deviations of each patch (Equation \ref{NCC}). However, as mentioned in previous sections, standard deviation calculation requires expensive square-root operations. So as to make the correlation calculation simpler on FPGA, platform, for some experiments we use mean-absolute-deviation (MAD) instead of standard-deviation.
\end{enumerate} 
\item \emph{Unnormalized Correlation}: As discussed in Section 2.2, the reason we employ normalized-cross correlation, instead of convolution, which the standard operation of a CNN, is simply to avoid over-fitting certain characteristics of infrared imagery. In order to experimentally portray this condition, using the same dataset, we train a two layer plain CNN with 4 filters. This CNN is the unnormalized version of the proposed NCC-NN structure. The results of this detection method is also presented in our results, namely as \emph{unnormalized correlation}.
\item \emph{Infrared Patch-Image (IPI) Model}: We have also chosen to add a method that is designed for infrared small-target detection. We believe that \emph{Infrared Patch-Image Model} \cite{gao2013infrared} is a representative technique for this purpose. It is based on the non-local self-correlation property of IR images and small target detection task is transformed into an optimization problem of recovering low-rank and sparse matrices.
\end{enumerate} 

\subsection{Detection Performance}
In this subsection we present the detection performance of the benchmarked methods. As mentioned in the previous subsection, for the Gaussian filter and \emph{The Hat} filter, different versions are also implemented. 

We have included a group of Gaussian filters with different standard deviations, 0.5, 1.2 ans 2.0 pixels, because these three versions showed the best performance on the given dataset. 

For \emph{The Hat} filter, we have implemented several versions, aiming at correctly evaluating the possible optimizations that can be applied on an FPGA platform. In addition to the ideal 15$ x $15 filter (where ideal refers to the unlimited precision of the filter values, i.e. MATLAB double precision\footnote{Gaussian and the unnormalized correlation filters are also ideal in precision.}), 9$ x $9 and 7$ x $7 cropped versions of the ideal filter are also included. In addition, 7$ x $7 and 5$ x $5 cropped versions are also implemented with fixed-point precision of 8bits. Moreover, as explained in the previous subsection, MAD-based correlation methods is also implemented for 9$ x $9 ideal, 9$ x $9 fixed, 7$ x $7 fixed-point and 5$ x $5 fixed-point versions.

\begin{figure*}[t]
\centering
\includegraphics*[trim=0 0 0 0,clip=true,width=1\textwidth]{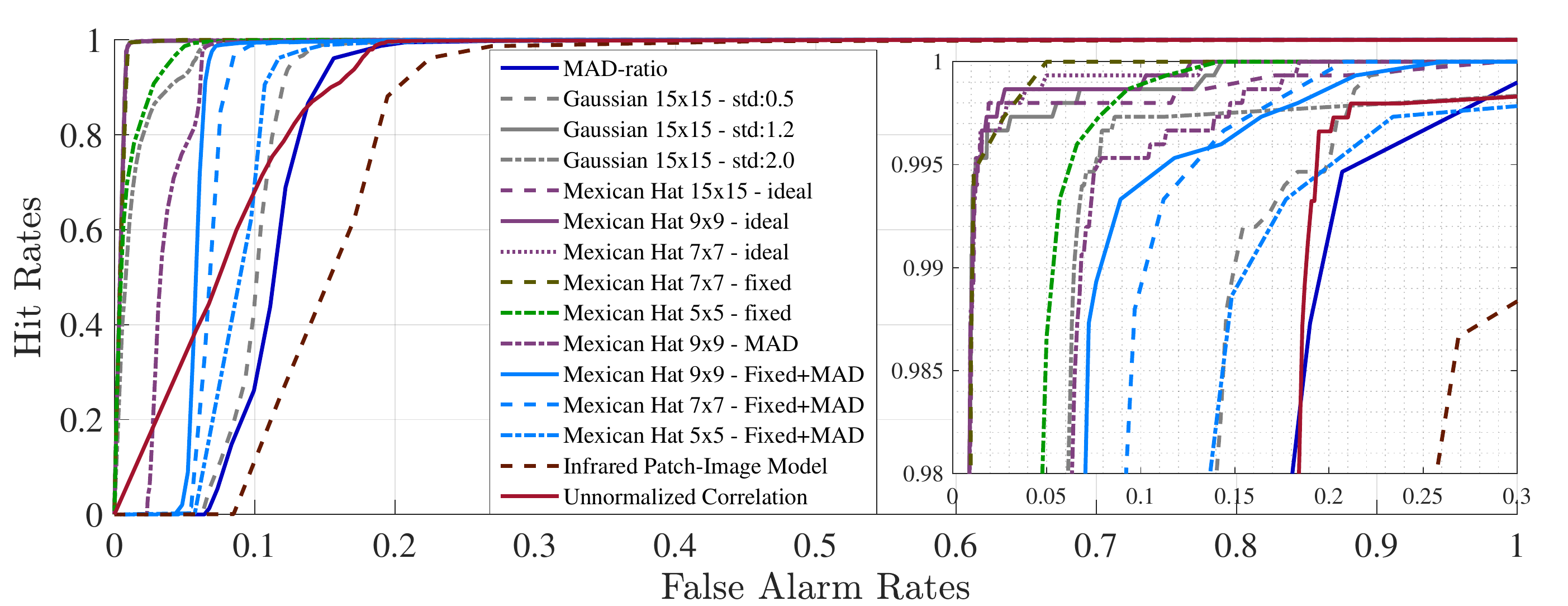} 
\caption{ROC curves for all experiments are depicted. The behaviors near HR=1 for all curves are depicted within the zoomed window.}
 \label{ROCResults}
\end{figure*}

In Figure \ref{ROCResults}, the ROC curves for each implemented methods is depicted. As it can be seen from the figure, while MAD-ratio method performs the worst, "The Hat" filters with ideal precision perform the best. The 15$ x $15 and 9$ x $9 ideal precision "The Hat" filters give the best results, whereas the performance considerably falls when the filter size is 7$ x $7 or lower. The Gaussian filter with 1.2 pixels standard deviation is the third best, with slightly worse false alarm rates. This is an expected result because the advantage of "The Hat" filter against the Gaussian filter is its ability to eliminate clutter. 

\begin{figure*}[t]
\centering
\begin{subfigure}{17cm}	
    \centering
	\includegraphics*[trim=0 30 0 0,clip=true,width=1\textwidth]{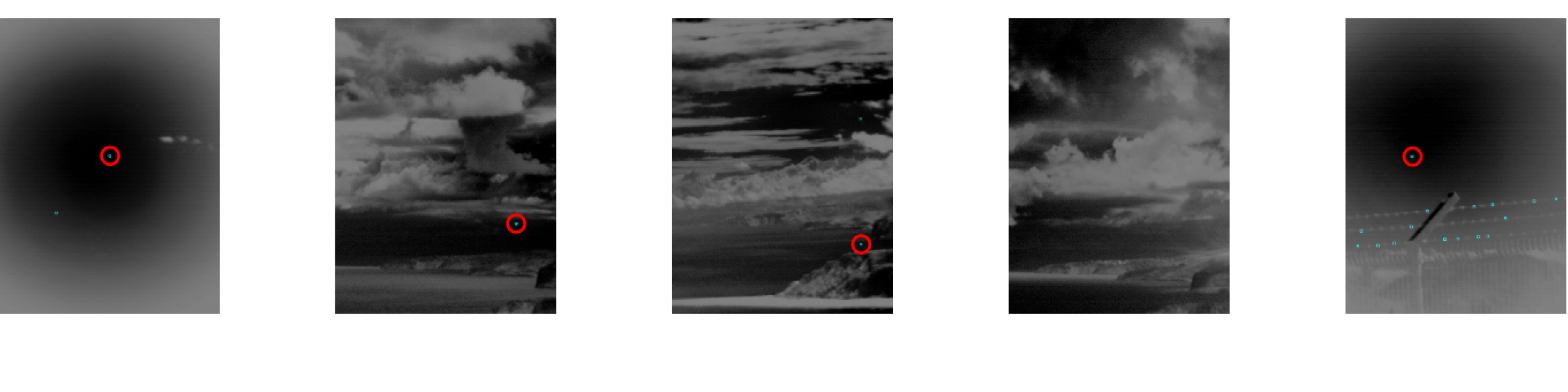}
	\caption{Results of MexHat 9x9 fixed-MAD thr=20}	
\end{subfigure}
\begin{subfigure}{17cm}	
    \centering
	\includegraphics*[trim=0 30 0 0,clip=true,width=1\textwidth]{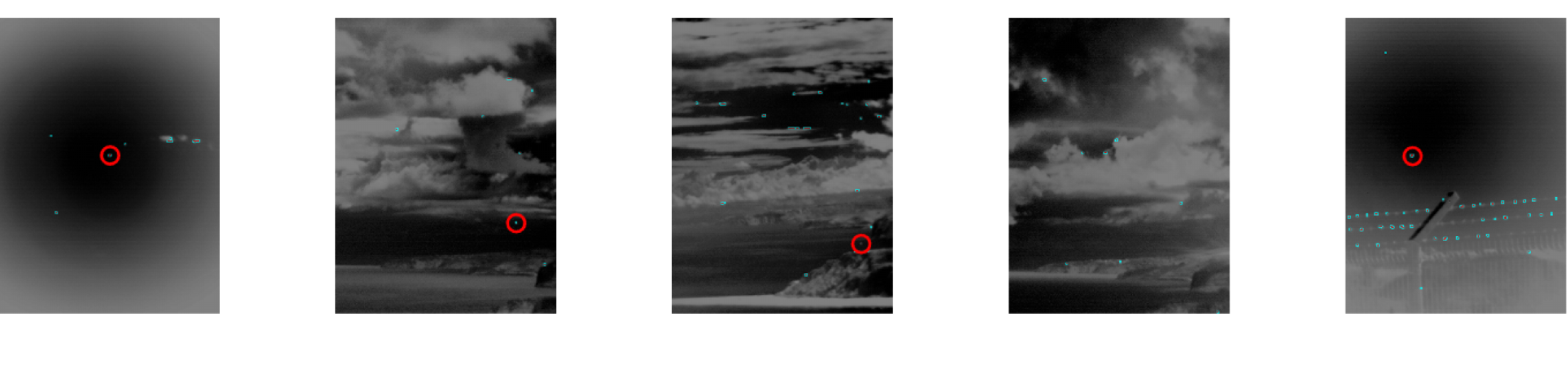}
	\caption{Results of MAD-ratio method}	
\end{subfigure}
\caption{Comparative results for 5 different scenes with and without targets for two methods are depicted. In each image the target are designated by red circles, while the fourth image from left has no target. Each declaration is shown with a cyan rectangle. Thus, cyan rectangles, not covered by a red circle are false alarms. In (a) results of the The Hat 9x9 MAD filter is given, while in (b) results of the MAD-ratio method is provided.}
 \label{ResultsFigure}
\end{figure*}

Although Figure \ref{ROCResults} shows that ideal precision is crucial for the best performance, it is usually not feasible for an FPGA platform. Therefore, we also search for an optimum filter with proper FPGA optimizations such as fixed-point precision and MAD-based correlation method.  In Figure \ref{ROCResults}, we can see that 9$ x $9, fixed-point and MAD-based implementation of "The Hat" filter (solid blue ROC curve) provides an average performance within the benchmarked methods. This implementation is the most feasible one considering filter size, FPGA size and run-time speed, for our FGPA-based system. Though this is not a generalization and may change for different systems and designs. 

Another important remark of the results is the high false alarm rates for the \emph{unnormalized correlation} and \emph{IPI} methods. As expected, the varying characteristics regarding the infrared dynamic range of different scenes increases the false alarm rates. We believe that, these methods' performances fall behind even the performances of simple filters like Gaussian, or MAD-ratio, mainly because of this phenomenon. 

In Figure \ref{ResultsFigure}, comparative results for five different scenes with and without targets for two methods, namely 9$ x $9, fixed-point and MAD-based \emph{The Hat filter} (solid blue ROC curve in Figure \ref{ROCResults}) and MAD-ratio method (solid dark-blue ROC curve in Figure \ref{ROCResults}) are depicted. In each image the targets are designated by red circles, while the fourth image from left has no target, but clutter. Each declaration is shown with a cyan rectangle. Thus, cyan rectangles, not covered by a red circle are false alarms. In Figure \ref{ResultsFigure}a, the results of "The Hat+9x9+MAD" filter is given, while in Figure \ref{ResultsFigure}b results of the MAD-ratio method is provided. We can see that, although optimized for an FPGA platform, in terms of precision, size and mathematical operations, ``The Hat'' filter is highly resilient against different types of clutter, such as terrain, cloud, bad-pixels, etc. 

\section{Conclusions}
We present a machine learning approach to the problem of infrared small target detection filter design. For this purpose, similarly to a convolutional layer of a neural network, the normalized-cross-correlational (NCC) layer is proposed. This is a modified version of the convollutional layer of a neural network, in which both the input patch and the filter is normalized, so that the convolution operation becomes identical to a correlation operation, if the filter is real and symmetric.

We work on a mid-wave band infrared small target dataset and train filters for detect small target and eliminate various type of clutter. Our benchmarking with different filters and a plain CNN show that for our dataset, a single filter with a special shape (which we call "The Hat") is sufficient to detect small targets and eliminate various clutter.

It is probable that this shape is only useful considering our dataset, thus we can only utilize it for this IR band and a similar optics. However we present a general framework to create this filter. That's why, for another dataset, the method can be applied and required filters can be obtained. As a future direction, we plan to obtain an extended target detection method and are working on creating a suitable dataset for this purpose. 

\begin{table}[t]
\centering
\small
\begin{tabular}{|l|c|c|c|c|c|}
\hline
\emph{algorithm:} & {MAD-Ratio} & {NCC with std.} & {NCC with mad.} & {IPI model} & {unnorm. corr.}\\ \hline
\hline
\emph{no. of mul.: } & N-sqr & N-sqr & N-sqr & k1 x N-sqr  & N-sqr \\
\emph{no. of add.: } & {N-sqr/f-sqr} & 2 x {N-sqr/f-sqr} & 2 x {N-sqr/f-sqr} & k3 x N-sqr  & {N-sqr/f-sqr} \\
\emph{no. of div.: } & {N-sqr/f-sqr} & {N-sqr/f-sqr} & {N-sqr/f-sqr} & k3 x N-sqr  & - \\
\emph{no. of sqr. root: } & {-} & {N-sqr/f-sqr}  & {-} & k4 x N-sqr  & {-}\\
\hline
\end{tabular}
\caption{Numbers of arithmetic operations for each bench-marked method, for an N-by-N image and a f-by-f filter size, are provided. IPI model \cite{gao2013infrared} includes an convergence loop, and number of operations depend on the context of the image. The number of operations are indicated with a modifier $k_{i}$, where  $k_{i}>1$.} 
\label{complexity}
\end{table}

\section*{Acknowledgements}
We would like to thank ASELSAN engineers Erkan Okuyan and Meri\c{c} K. Karakurt for sharing their ideas and experience on how to optimize our algorithms on an FPGA platform.

%\bibliographystyle{elk}
%\bibliography{references}

\end{document}